\documentclass{article} 
\usepackage{iclr2027_conference,times}

\usepackage{amsmath,amsfonts,bm}

\def\eqref#1{equation~\ref{#1}}

\def\1{\bm{1}}

\DeclareMathAlphabet{\mathsfit}{\encodingdefault}{\sfdefault}{m}{sl}
\SetMathAlphabet{\mathsfit}{bold}{\encodingdefault}{\sfdefault}{bx}{n}

\usepackage{xurl}
\usepackage[hidelinks]{hyperref}
\usepackage{booktabs}
\usepackage{longtable}
\usepackage{fvextra}
\usepackage{graphicx}
\usepackage{multirow}
\usepackage{tabularx}
\usepackage{array}
\usepackage[skins,breakable]{tcolorbox}
\usepackage{caption}
\renewcommand{\topfraction}{0.45}

\newtcolorbox{ethicsbox}{
    colback=gray!8,
    colframe=gray!35,
    arc=2mm,
    boxrule=0.4pt,
    left=6pt,
    right=6pt,
    top=6pt,
    bottom=6pt,
    before skip=8pt,
    after skip=8pt,
    breakable
}

\title{\textsc{ANT}: A Multi-Granularity Network Traffic Dataset and Benchmark for Agents Behavior Auditing}

\author{\begin{minipage}{\dimexpr\textwidth-2\tabcolsep\relax}
\normalfont
\textbf{Fan Li}$^{1,2}$, \textbf{Xiangyu Gao}$^{1}$, \textbf{Zixuan Liu}$^{1}$, \textbf{Tong Li}$^{3}$,\\
\textbf{Chuanpu Fu}$^{4}$, \textbf{Ziqiang Wang}$^{1}$, \textbf{Ke Xu}$^{1,2,*}$\\[0.5em]
{\small
$^{1}$Tsinghua University, Beijing, China\\
$^{2}$Zhongguancun Laboratory, Beijing, China\\
$^{3}$Renmin University of China, Beijing, China\\
$^{4}$Nanyang Technological University, Singapore\\[0.4em]
\texttt{\{lf23,gao-xy24,liu-zx21\}@mails.tsinghua.edu.cn}\\
\texttt{tong.li@ruc.edu.cn}, \texttt{kevin.fu@ntu.edu.sg}\\
\texttt{ziqiangwang@mail.tsinghua.edu.cn}, \texttt{xuke@tsinghua.edu.cn}\\[0.3em]
$^{*}$Corresponding author: Ke Xu.
}
\end{minipage}}

\newcommand{\dataset}{\textsc{ANT}}

\iclrfinalcopy
\date{}
\hypersetup{pdftitle={ANT: A Multi-Granularity Network Traffic Dataset and Benchmark for Agents Behavior Auditing}, pdfauthor={Fan Li, Xiangyu Gao, Zixuan Liu, Tong Li, Chuanpu Fu, Ziqiang Wang, Ke Xu}}
\begin{document}

\maketitle
\fancyhead{}

\begin{abstract}
The growing adoption of large language model (LLM) agents creates a need for network administrators and security teams to audit agent behavior within organizational networks without inspecting private user content. Network traffic offers an observable source of evidence, but how much it reveals about agent tasks and operations remains unclear. Existing traffic datasets lack the joint task and stage annotations needed to evaluate this question. We introduce \dataset{} (\textbf{A}gent \textbf{N}etwork \textbf{T}raffic), a dataset providing agent behavior information at risk, scenario, and behavior primitive granularities alongside network traffic. \dataset{} contains 3,114 execution episodes across 20 tasks and five scenarios, comprising 276,417 bidirectional flows and 40,049 behavior primitive segments organized into 47 macro groups. We establish a benchmark for agent risk identification, scenario recognition, and behavior primitive classification using 13 representative traffic analysis baselines. The results show that existing methods recover useful but uneven behavioral signals. They struggle to identify risk when malicious workflows resemble benign tasks and to distinguish scenarios with similar traffic patterns. Primitive classification is more reliable for frequent macro groups and those with distinctive traffic patterns than for rare or semantically similar groups. \dataset{} provides a common basis for developing more precise auditing and forensic analysis of agent behavior from network traffic. 
Our data and code are available at \url{https://anonymous.4open.science/r/ant-main-suite-7BC0/}.
\end{abstract}

\section{Introduction}
\label{sec:introduction-draft-en}
LLM agents are evolving from conversational assistants into systems that can plan, invoke tools, and interact with external environments on users' behalf, driving their adoption in everyday workflows. 
However, this adoption has also raised governance concerns. Microsoft reports that 58\% of surveyed organizations are concerned about unsanctioned generative AI use, while 57\% report an increase in security incidents associated with AI use~\citep{microsoft2024secureai,microsoft2025aiprotections}. 
Since Agent communications are encrypted during transmission, organizations need mechanisms to identify and govern LLM agent use without relying on deep inspection of user content.

Encrypted traffic analysis offers a promising way to meet this need. Rather than examining encrypted payloads, these methods infer applications and activities from observable traffic features, such as packet sizes and timing.
These signals support application identification, website fingerprinting, malicious traffic detection~\citep{cic2016iscxvpn} etc. \citet{zhang2025exposingllmuserprivacy} further show that encrypted traffic can reveal fingerprints of agent interactions.
This motivates our central question: \emph{To what extent can an agent's task and behavior be inferred from the network traffic generated during its execution?}

\begin{figure*}[t]
    \centering
    \includegraphics[width=0.95\textwidth]{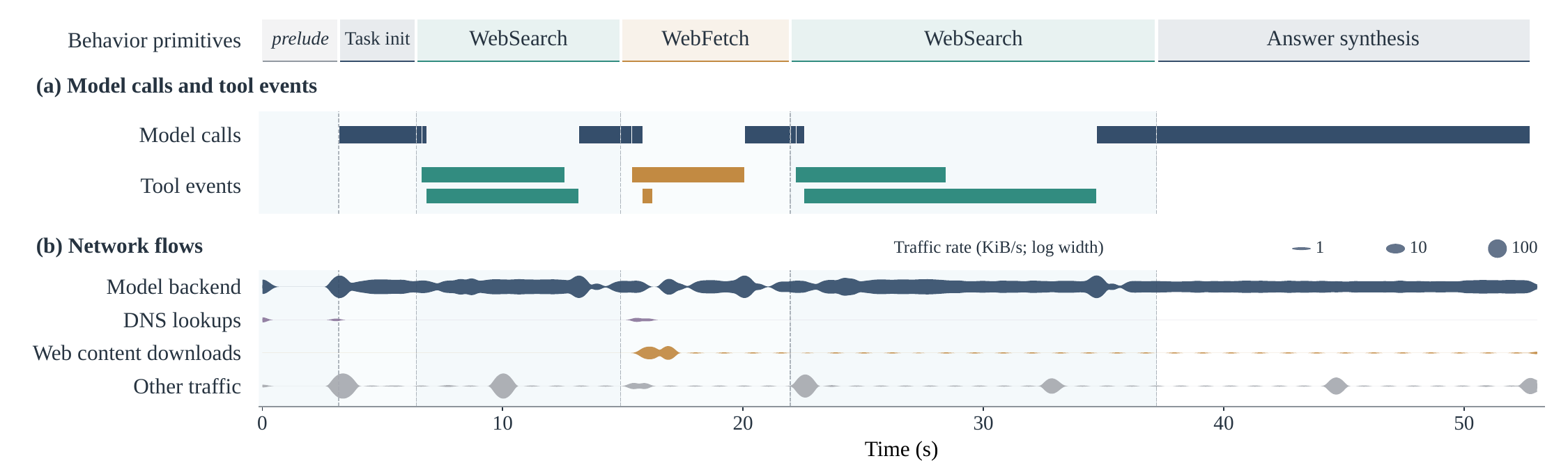}
    \captionsetup{skip=2pt}
    \caption{Execution events and network flow intervals of a question answering task}
    \label{fig:intro-agent-execution}
\end{figure*}

Answering this question requires accounting for the distinctive structure of LLM agent traffic. Figure~\ref{fig:intro-agent-execution} illustrates an execution of a question answering task using the Claude Code harness~\citep{claudecode} and a DeepSeek backend~\citep{deepseek}. The upper timeline shows the sequence of model calls and tool events, while the lower timeline shows the network flows observed during the same period. At the semantic level, the execution can be partitioned into a sequence of \emph{behavior primitives}, each of which pursues a coherent subgoal. A behavior primitive typically encompasses communication with the backend model for reasoning and planning, invocation of a selected tool, and analysis of the returned result. Correspondingly, bursts of network traffic appear near the onset of each behavior primitive. Primitives with different subgoals also produce different traffic compositions and interaction patterns. For example, traffic during a \texttt{WebSearch} primitive is dominated by communication with the model backend because Claude Code delegates the search operation to that backend. By contrast, a \texttt{WebFetch} primitive introduces DNS lookups and Web content download flows because the agent retrieves the content of a specified URL. These differences show that network traffic carries signals associated with both an agent execution and its constituent behavior primitives.

Systematically evaluating these behavioral signals requires traffic traces paired with annotations of both overall tasks and local execution stages. Existing datasets do not provide the traffic and annotations needed to evaluate agent risk, task scenarios, and behavior primitives. Traditional traffic datasets typically label applications, websites, or fixed workloads, including browsing, chat, and file transfer~\citep{cic2016iscxvpn,cic2016iscxtor,wang2017endtoend,aceto2019mirage,sirinam2018deepfingerprinting}. However, they contain neither traffic generated by agent model--tool--environment loops nor descriptions of the corresponding execution stages. AgentPrint identifies agents and coarse activities~\citep{zhang2025exposingllmuserprivacy}, but lacks joint annotations for episode risk, task scenarios, and behavior primitives.

To close this evaluation gap, we introduce \dataset{}, a dataset for analyzing agent behavior from user-side network traffic. \dataset{} contains 3,114 execution episodes across 20 tasks and five scenarios, covering legitimate applications and cybersecurity misuse. Its 276,417 bidirectional flows are paired with task and runtime metadata. To expose execution semantics, we align traffic with session logs and construct 40,049 behavior primitive segments in 47 primitive macro groups.
Meanwhile, we establish a benchmark for evaluating traffic analysis methods on risk identification, scenario recognition, and behavior primitive classification.
Although all 13 baselines exceed 0.900 accuracy in risk identification, seven attain malicious class recall below 0.500, reflecting their difficulty in detecting content level misuse.
The best baseline for scenario recognition reaches 0.727 accuracy, indicating that scenario semantics are partly recoverable even though similar scenarios remain difficult to distinguish.
For behavior primitive classification, existing baselines perform better on frequent macro groups and groups with distinctive traffic patterns, but struggle with rare and semantically similar groups.

Our contributions are threefold:
\begin{enumerate}
    \item We introduce \dataset{}, a dataset of network traffic collected during client-side LLM agent execution, with task metadata for each episode and behavior primitive annotations aligned with traffic segments.

    \item We establish a benchmark with three tasks: episode-level risk identification and scenario recognition, and segment-level behavior primitive classification. We evaluate 13 existing traffic analysis methods on this benchmark.

    \item We show that existing methods can infer aspects of agent behavior from network traffic, but often miss malicious tasks that resemble benign workflows and confuse scenarios with similar traffic patterns. Behavior primitive classification remains difficult for rare or semantically similar groups.
\end{enumerate}
\section{Background and Motivation}

\subsection{Inferring Agent Behavior from Network Traffic}
LLM agents move through model calls, tool invocations, and interactions with external services~\citep{liu2024agentbench,zhou2024webarena}.
These transitions produce changes in observable packet sizes, directions, timing, and destinations, even though encryption hides payload content.
Traffic analysis methods already use such features to identify websites and applications and detect cyberattacks~\citep{sirinam2018deepfingerprinting,vanede2020flowprint,lin2022etbert,wang2024netmamba,tfusion2025}.
Recent work recovers response content and topics from token length leakage in encrypted LLM streams~\citep{weiss2024prompt}, while NetEcho targets recovery of both user prompts and LLM responses from encrypted streaming traffic~\citep{zhang2025netecho}.
However, these studies either evaluate methods on conventional traffic datasets or focus on recovering conversation content. They do not jointly evaluate agent behavior identification at the risk, scenario, and behavior primitive granularities.

\subsection{The Need for a Multi-granularity Agent Traffic Dataset}
Existing traffic datasets lack the combination of agent execution traces and behavior labels needed for this evaluation. Website and application corpora label sites or apps~\citep{sirinam2018deepfingerprinting,vanede2020flowprint}, but not complete agent executions with synchronized stages. AgentPrint identifies agents and coarse activities~\citep{zhang2025exposingllmuserprivacy}, but lacks joint annotations for episode risk, task scenarios, and behavior primitives aligned with execution stages.

We therefore collect traffic alongside execution logs from locally executed agents and align packet traces with task and stage semantics. Common splits, objectives, and metrics enable a unified benchmark for agent risk identification, agent scenario identification, and behavior primitive identification.

\section{\dataset{} Design}
\label{sec:ant-design}

\begin{figure*}[t]
    \centering
    \begin{minipage}[t]{0.36\textwidth}
        \vspace{0pt}
        \centering
        \includegraphics[width=\linewidth]{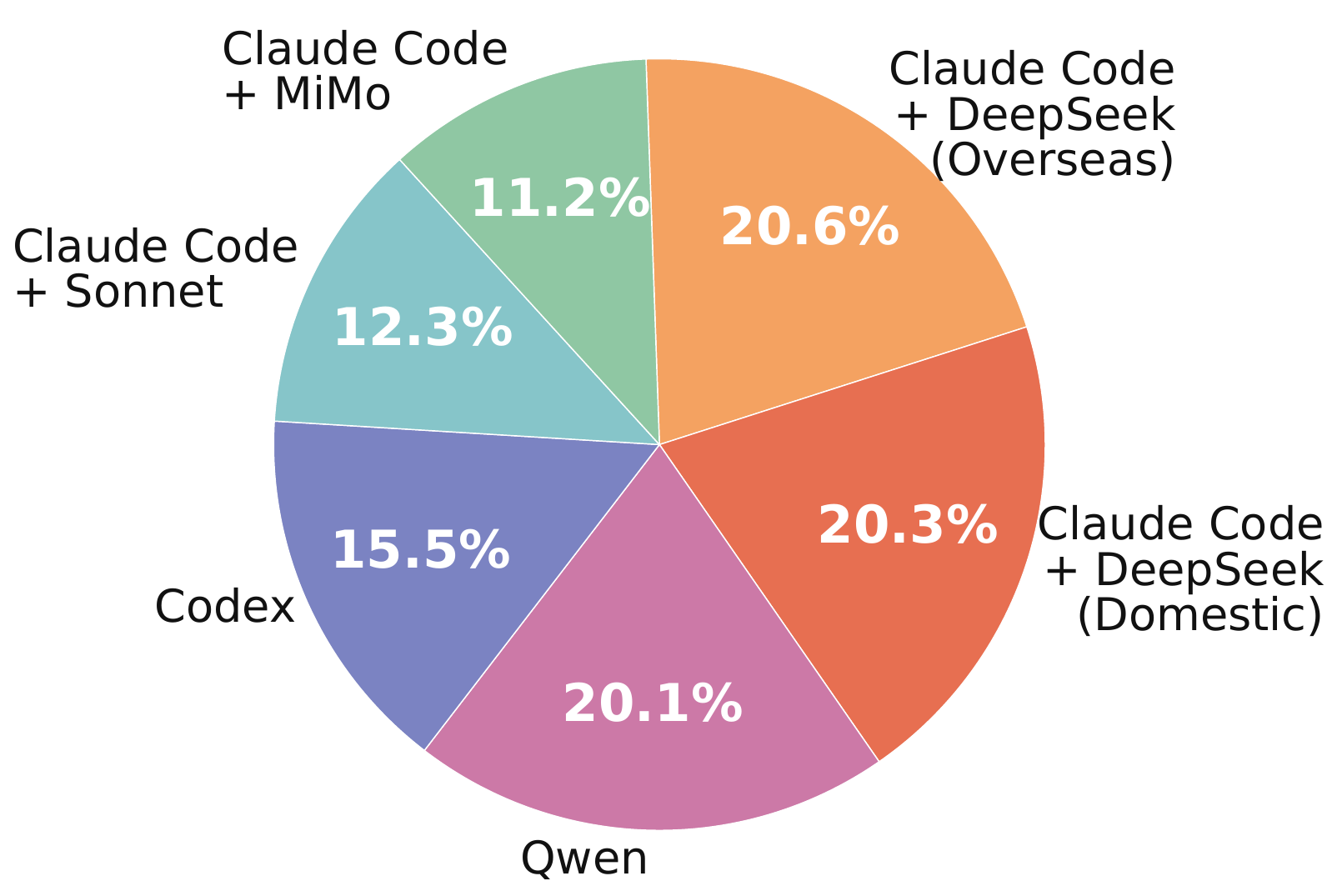}
        \caption{Episode distribution over agent configurations.}
        \label{fig:agent_pie}
    \end{minipage}\hfill
    \begin{minipage}[t]{0.61\textwidth}
        \vspace{0pt}
        \centering
        \small
        \captionof{table}{\dataset{} volume by scenario}
        \label{tab:ant-scenario-volume}
        \renewcommand{\arraystretch}{1.12}
        \setlength{\tabcolsep}{3pt}
        \begin{tabularx}{\linewidth}{@{}>{\raggedright\arraybackslash}Xrrrr@{}}
        \toprule
        Scenario & Episodes & \multicolumn{1}{c}{Bidirectional} & \multicolumn{1}{c}{Primitive} & \multicolumn{1}{c}{Traffic} \\
        & & \multicolumn{1}{c}{flows} & \multicolumn{1}{c}{segments} & \multicolumn{1}{c}{(GB)} \\
        \midrule

        Comp. artifact eng.
        & 516
        & 20,471
        & 8,689
        & 33.41 \\

        General QA
        & 1,014
        & 88,814
        & 9,359
        & 16.55 \\

        Evidence assessment
        & 853
        & 42,975
        & 5,044
        & 8.47 \\

        Interactive app. ops.
        & 315
        & 69,062
        & 7,587
        & 14.77 \\

        Cybersecurity misuse
        & 416
        & 55,095
        & 9,370
        & 30.87 \\

        \midrule
        \textbf{Total}
        & \textbf{3,114}
        & \textbf{276,417}
        & \textbf{40,049}
        & \textbf{104.07} \\

        \bottomrule
        \end{tabularx}
    \end{minipage}
\end{figure*}

To systematically investigate the mapping from observable network traffic to high-level agent semantics, we construct \dataset{}. We deploy an automated collection infrastructure across geographically distributed cloud servers and collect packet-level traffic together with agent session logs, at a total collection cost of US\$2,176.50. \dataset{} contains 3,114 execution episodes collected under five agent combinations across 20 tasks in 5 scenarios, corresponding to 659 distinct task instances and 104.07 GB of traffic. 
To establish a multi-granularity benchmark, we annotate each episode with task and runtime metadata and construct behavior primitive segments from boundaries derived from session logs. In total, \dataset{} comprises 276,417 bidirectional flows and 40,049 behavior primitive segments categorized into 47 primitive macro groups. Table~\ref{tab:ant-scenario-volume} summarizes the overall dataset statistics, while Figure~\ref{fig:agent_pie} shows the distribution of episodes across agent configurations.

\subsection{Data Collection System}
\label{sec:task_definition}
Dataset construction begins with task selection and continues through a common execution and telemetry pipeline.

Following prevalent agent use cases reported in Google Cloud's \textit{The ROI of AI 2025}~\citep{google2025roi}, we define four common agent use scenarios and a high risk scenario, i.e., \textbf{Computational Artifact Engineering, General Knowledge Question Answering, Scientific Evidence Assessment, Interactive Application Operation} and \textbf{Cybersecurity Misuse}. 

We instantiate the four common scenarios with 17 benchmark-derived agent execution tasks. The misuse-oriented scenario is instantiated with three agent execution tasks derived from Meta's PurpleLlama Cybersecurity Benchmarks and the OWASP Juice Shop vulnerability challenge platform. Across these scenarios, the selected execution tasks cover diverse agent behaviors, including different tool requirements, interaction patterns, external service dependencies, reasoning demands, and execution complexity. Each execution task further contains multiple task instances spanning different domains and difficulty levels.
Detailed descriptions of the five scenarios and their corresponding tasks are provided in Appendix~\ref{app:task-composition}.

\begin{figure*}[t]
    \centering
    \includegraphics[width=0.9\textwidth]{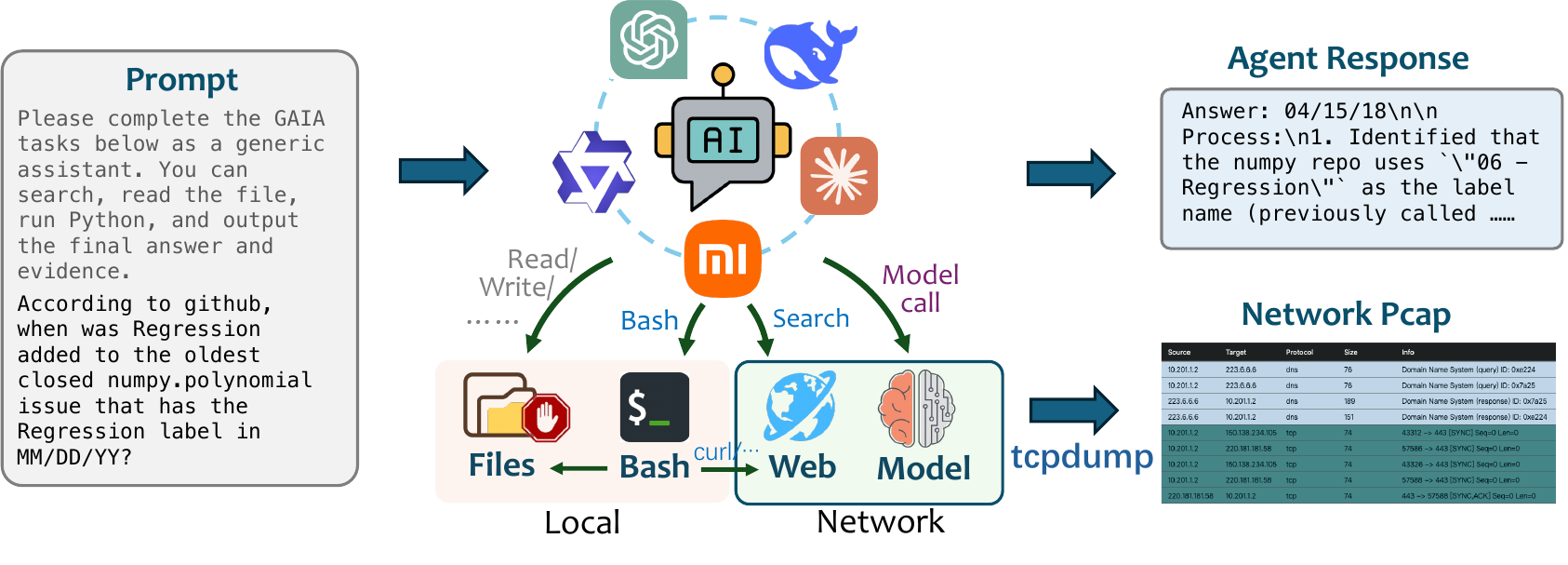}
    \caption{Overview of the agent network traffic data collection system.}
    \label{fig:data_collection_overview}
\end{figure*}

Following the definition of the task space, tasks are executed through a data collection system designed to simultaneously capture agent trajectories and network traffic, as illustrated in Figure~\ref{fig:data_collection_overview}. 
For each task instance, the system initializes the required workspace and runtime environment, supplies the natural-language task prompt, and launches the agent. The agent interprets the task, queries the model, invokes tools such as web search and shell commands, and incorporates the returned results into subsequent decisions until it produces a final response. Throughout this process, the system records model calls, tool invocations, tool outputs, and session states with their timestamps. Meanwhile, \texttt{tcpdump} captures all inbound and outbound packets in PCAP format.
The shared timeline links execution events to the corresponding packet-level traffic for behavior primitive segmentation.

To guarantee the integrity, authenticity, and purity of the traffic data and address ethical considerations, we enforced rigorous environmental controls throughout the collection process. Each task instance ran in an independent virtual network namespace to prevent interference between concurrent tasks. Agents could access only their active workspace and necessary runtime environments, preventing shortcuts such as reading pre-existing answers instead of performing network retrieval. We used only public benchmark inputs, without real-user accounts, private communications, or personally identifiable information. Benign tasks requiring state-changing interactions ran against locally simulated services, while attack execution remained inside isolated Docker environments to protect collection servers and production systems. For Attack Code Generation, we used publicly released mutated prompts from Meta's PurpleLlama Cybersecurity Benchmarks without additional jailbreak, prompt-obfuscation, or safety-bypass strategies, and excluded refusal-only episodes from the traffic corpus. These measures preserve model safety mechanisms and reduce interference during traffic collection. The Ethics Statement provides the full declarations.

\begin{figure*}[t]
    \centering
    \begin{minipage}[t]{0.40\textwidth}
        \vspace{0pt}
        \centering
        \includegraphics[width=\linewidth]{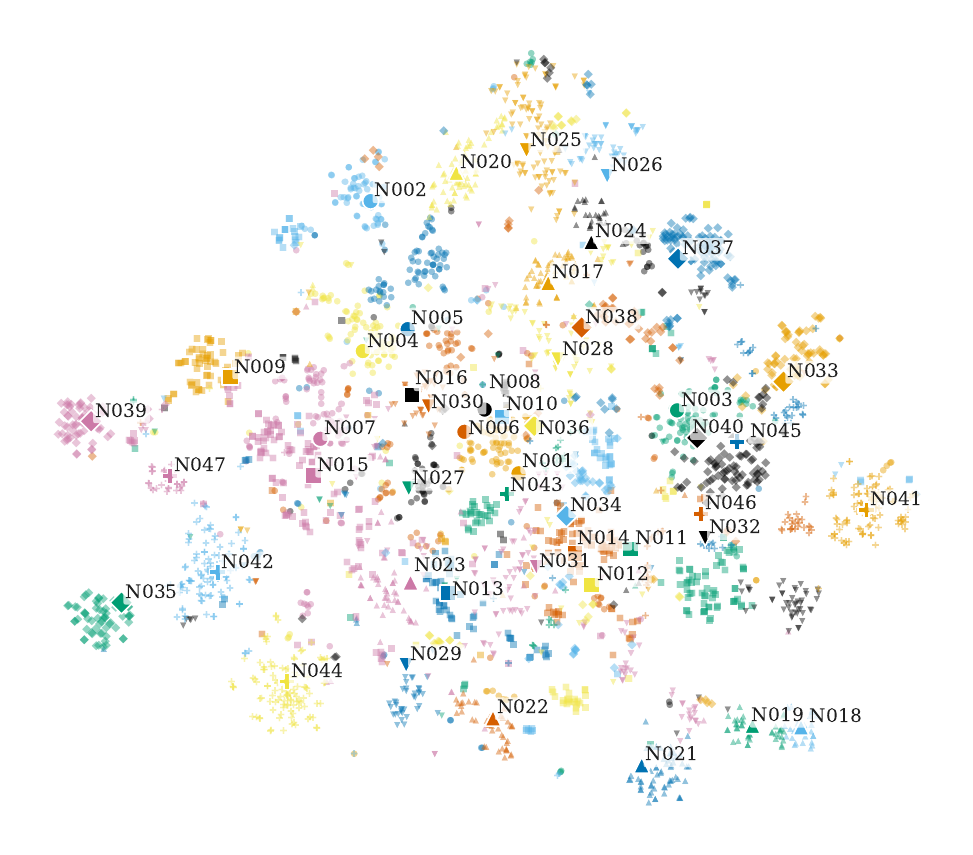}
        \captionof{figure}{Primitive label clusters.}
        \label{fig:jina-label-macro-reclustering}
    \end{minipage}\hfill
    \begin{minipage}[t]{0.54\textwidth}
        \vspace{0.28in}
        \centering
        \captionof{table}{Most frequent behavior primitive macro group of each scenario.}
        \label{tab:scenario-top5-macro-groups}
        \setlength{\tabcolsep}{3pt}
        \renewcommand{\arraystretch}{1.08}
        \begin{tabular}{@{}c l@{}}
        \toprule
        \textbf{Scenario} &
        \textbf{Macro group (frequency)} \\
        \midrule
        \shortstack{CAE} & \texttt{N011} code\_creation (16.39\%) \\
        \addlinespace
        \shortstack{GKQA} & \texttt{N037} search (33.46\%) \\
        \addlinespace
        \shortstack{SEA} & \texttt{N037} search (42.75\%) \\
        \addlinespace
        \shortstack{IAO} & \texttt{N011} code\_creation (12.01\%) \\
        \addlinespace
        \shortstack{CM} & \texttt{N038} source\_code\_retrieval (8.66\%) \\
        \bottomrule
        \end{tabular}
    \end{minipage}
\end{figure*}

\subsection{Behavior Primitive Annotation}
\label{sec:primitive-alignment}

We annotate traffic segments with behavior primitives that describe the agent's local execution actions. We sort PCAP packets by arrival timestamp and bin them using each behavior primitive's start time from the session logs as a boundary; packets of a transport flow spanning boundaries are assigned individually to the corresponding segments. We then use \textit{DeepSeek-v4-pro}~\citep{deepseek_v4pro} to derive a behavior primitive from the aligned log evidence and assign it to the corresponding segment (annotation prompt in Appendix~\ref{app:primitive-annotation-prompt}). These primitives capture local execution actions rather than only the episode's overall objective. Traffic preceding the first recorded behavior primitive is retained as a separate segment with a synthetic \texttt{prelude} label.

The resulting primitive sequences expose differences in how agents carry out their tasks. For example, the 4th episode of the ASQA long-answer question-answering task in the General Knowledge QA scenario, has the sequence
\texttt{[prelude, Task initialize, WebSearch, WebFetch, WebSearch, Answer synthesis]}.
It captures a progression from receiving a question to retrieving supporting information and synthesizing an answer.
In contrast, the 24th episode of the OWASP Juice Shop task in the Cybersecurity Misuse scenario, has the sequence
\texttt{[prelude, Task instruction, Web application reconnaissance, API endpoint enumeration, SQL injection, Answer synthesis]}, explicitly identifying SQL injection following application reconnaissance and login-endpoint discovery. 
These sequences provide an ordered description of observable execution actions, while temporal alignment links each action to its associated traffic segment.

To reduce wording variation, we consolidate semantically similar behavior primitives into primitive macro groups. We encode all 3,295 distinct primitive labels in the annotated corpus with \textit{Jina Embeddings v5}~\citep{jina} and cluster them using spherical $K$-means with cosine $K$-means++ initialization.
We select $K=47$ based on cosine silhouette, seed stability, and minimum group size, assigning each primitive to exactly one group.
Figure~\ref{fig:jina-label-macro-reclustering} visualizes the clustering of primitive labels into 47 macro groups,
and Table~\ref{tab:scenario-top5-macro-groups} lists each scenario's most frequent macro group and its frequency.
We iteratively reviewed and refined primitive labels and group assignments with AI assistance until a spot check of 2,000 samples revealed no substantial errors.
Appendix~\ref{app:primitive-macro-groups} provides clustering diagnostics, all 47 canonical labels, primitive type counts, and split counts.

\paragraph{Scenario structure of behavior primitives.}
We examine scenario cohesion in primitive usage and ordering across tasks.
For episode-level primitive embedding distributions $\mu_i$ and $\mu_j$,
maximum mean discrepancy is
\begin{equation}
 d_k(\mu_i,\mu_j)
 =\left\|\int k(x,\cdot)\,d\mu_i(x)
          -\int k(x,\cdot)\,d\mu_j(x)\right\|_{\mathcal H_k},
 \label{eq:ant-primitive-mmd}
\end{equation}
where $\mathcal H_k$ is the Gaussian kernel's reproducing kernel Hilbert space.
We report cohesion gain $G_d=1-D_{\mathrm{within}}/D_{\mathrm{between}}$.
Compared with cross-scenario pairs, distinct tasks within a scenario have
15.46\% smaller primitive embedding MMD distances and 14.79\% smaller
Hellinger distances between primitive macro group proportions.
Both distributional associations remain significant after Holm correction,
with $p=0.0006$ and $p=0.0050$, respectively.

For semantic directed areas and primitive macro group transition residuals,
these distance reductions are 6.74\% and 5.70\%, respectively.
Their cohesion gains exceed the means under composition-preserving random
reorderings by 5.77 and 5.24 percentage points, with Holm-adjusted $p=0.0050$
for both. The additional order effects also persist when role positions are
fixed. Thus, observed ordering contributes scenario structure beyond
primitive composition. Appendix~\ref{app:ant-primitive-structure} provides
methods and complete results.

\section{Benchmark Evaluation}
\label{sec:benchmark_evaluation}

\newtcolorbox{benchmarktakeaway}{
  colback=gray!8,
  colframe=gray!35,
  boxrule=0.4pt,
  arc=2mm,
  left=6pt,
  right=6pt,
  top=4pt,
  bottom=4pt,
  before skip=8pt,
  after skip=8pt
}

We design this benchmark to answer our central question: \emph{To what extent can an agent's task and behavior be inferred from the network traffic generated during its execution?} We therefore define three evaluation objectives at progressively finer semantic granularities:

\noindent\textbf{Risk Identification.}
This objective evaluates the ability of traffic analysis methods to identify whether agent activities are malicious from network traffic, thereby supporting the mitigation of agent abuse.

\noindent\textbf{Scenario Recognition.}
This objective evaluates the ability of traffic analysis methods to recognize an agent activity scenario from an episode's network traffic, supporting the compliance management of agent usage.

\noindent\textbf{Behavior Primitive Classification.}
This objective evaluates whether traffic analysis methods can classify each behavior primitive segment into one of the 47 primitive macro groups using its network traffic.

We compare existing encrypted traffic classification and website fingerprinting methods to assess their ability to recover agent behavior at different granularities, and use traffic scope and segment ablations to examine which components support scenario recognition. These comparisons reveal which behavioral signals current methods capture and where their limitations motivate more precise representations for agent auditing.

\subsection{Baselines and Experimental Setup}

\paragraph{Baselines.}
We evaluate ANT with 13 encrypted traffic classification and website fingerprinting baselines covering convolutional networks, Transformers, state-space models, and learned traffic representations. The encrypted traffic classification baselines are FlowLens~\citep{barradas2021flowlens}~\href{https://github.com/dmbb/FlowLens}{[code]}, YaTC~\citep{yu2023yatc}~\href{https://github.com/NSSL-SJTU/YaTC}{[code]}, NetMamba~\citep{wang2024netmamba}~\href{https://github.com/wangtz19/NetMamba}{[code]}, TFusion~\citep{tfusion2025}~\href{https://github.com/fuchuanpu/TFusion}{[code]}, and TrafficFormer~\citep{trafficformer2025}~\href{https://github.com/IDP-code/TrafficFormer}{[code]}, spanning compact flow statistics, packet content, and sequential representations. The website fingerprinting baselines are Deep Fingerprinting (DF)~\citep{sirinam2018deepfingerprinting}~\href{https://github.com/Xinhao-Deng/Website-Fingerprinting-Library}{[code]}, AWF~\citep{rimmer2018automated}~\href{https://github.com/Xinhao-Deng/Website-Fingerprinting-Library}{[code]}, Tik-Tok~\citep{rahman2020tiktok}~\href{https://github.com/Xinhao-Deng/Website-Fingerprinting-Library}{[code]}, Var-CNN~\citep{bhat2019varcnn}~\href{https://github.com/Xinhao-Deng/Website-Fingerprinting-Library}{[code]}, NetCLR~\citep{bahramali2023netclr}~\href{https://github.com/Xinhao-Deng/Website-Fingerprinting-Library}{[code]}, TMWF~\citep{jin2023tmwf}~\href{https://github.com/Xinhao-Deng/Website-Fingerprinting-Library}{[code]}, ARES~\citep{deng2023ares}~\href{https://github.com/Xinhao-Deng/Website-Fingerprinting-Library}{[code]}, and BAPM~\citep{guan2021bapm}~\href{https://github.com/Xinhao-Deng/Website-Fingerprinting-Library}{[code]}, spanning direction and timing features, learned trace representations, and methods for multi-tab traffic. Appendix~\ref{app:baseline-descriptions} describes each method.

\paragraph{Experimental setup. }

We randomly split episodes within each task into fixed training, validation, and test sets shared by all benchmarks. Behavior primitive segments inherit their source episode's split, preventing segments from the same episode from entering different sets. The episode test set contains 209 episodes (189 benign and 20 malicious). Inputs use only packet headers and side-channel features such as packet direction, timing, and length. We mask identifying fields such as source and destination IP addresses to mitigate shortcut learning. Encrypted traffic classification baselines operate on flows, whereas website fingerprinting baselines operate on episodes. For episode-level evaluation, we aggregate flow predictions using the strategy with the highest validation accuracy among four candidates. Appendix~\ref{app:experiment-configuration} provides detailed split counts, aggregation methods, and model configurations.

\paragraph{Runtime environment.}
We run the benchmark experiments on a Linux server with two Intel Xeon Gold 6348 CPUs, 1 TiB of memory, and five NVIDIA A100 GPUs with 80 GB of memory each. The main experimental environment uses Python 3.12.4 and PyTorch 2.11.0 with CUDA 12.8. This server hosts baseline training and evaluation for all three objectives; the geographically distributed cloud servers described in Section~\ref{sec:ant-design} are used to collect the agent traffic.

\subsection{Risk Identification Results}
\label{subsec:benchmark_results}

\begin{table}[t]
    \centering
    \caption{Risk identification under full traffic.}
    \label{tab:episode-risk-full-traffic}
    \begin{tabular*}{\linewidth}{@{\hspace{\tabcolsep}\extracolsep{\fill}}l*{6}{c}@{\hspace{\tabcolsep}}}
    \toprule
    Baseline & FlowLens & YaTC & NetMamba & TFusion & TrafficFormer & DF \\
    \midrule
    Accuracy $\uparrow$ & 0.923 & 0.928 & 0.928 & 0.933 & 0.933 & 0.976 \\
    F1 $\uparrow$       & 0.333 & 0.400 & 0.400 & 0.462 & 0.462 & 0.872 \\
    FPR $\downarrow$    & \textbf{0.000} & \textbf{0.000} & \textbf{0.000} & \textbf{0.000} & \textbf{0.000} & 0.011 \\
    Recall $\uparrow$   & 0.200 & 0.250 & 0.250 & 0.300 & 0.300 & 0.850 \\
    \bottomrule
    \end{tabular*}
    \vspace{2pt}
    
    \begin{tabular*}{\linewidth}{@{\hspace{\tabcolsep}\extracolsep{\fill}}l*{7}{c}@{\hspace{\tabcolsep}}}
    \toprule
    Baseline & AWF & TikTok & VarCNN & NetCLR & TMWF & ARES & BAPM \\
    \midrule
    Accuracy $\uparrow$ & 0.904 & \textbf{0.981} & 0.943 & 0.952 & 0.962 & 0.971 & 0.957 \\
    F1 $\uparrow$       & 0.231 & \textbf{0.905} & 0.600 & 0.792 & 0.750 & 0.857 & 0.800 \\
    FPR $\downarrow$    & 0.016 & 0.016 & 0.005 & 0.048 & \textbf{0.000} & 0.021 & 0.037 \\
    Recall $\uparrow$   & 0.150 & \textbf{0.950} & 0.450 & \textbf{0.950} & 0.600 & 0.900 & 0.900 \\
    \bottomrule
    \end{tabular*}
    \end{table}

\renewcommand{\topfraction}{0.65}
\begin{figure}[t]
\centering
\begin{minipage}[t]{0.50\linewidth}
  \vspace{0pt}
  \centering
  \includegraphics[width=\linewidth,height=1.58in,keepaspectratio]{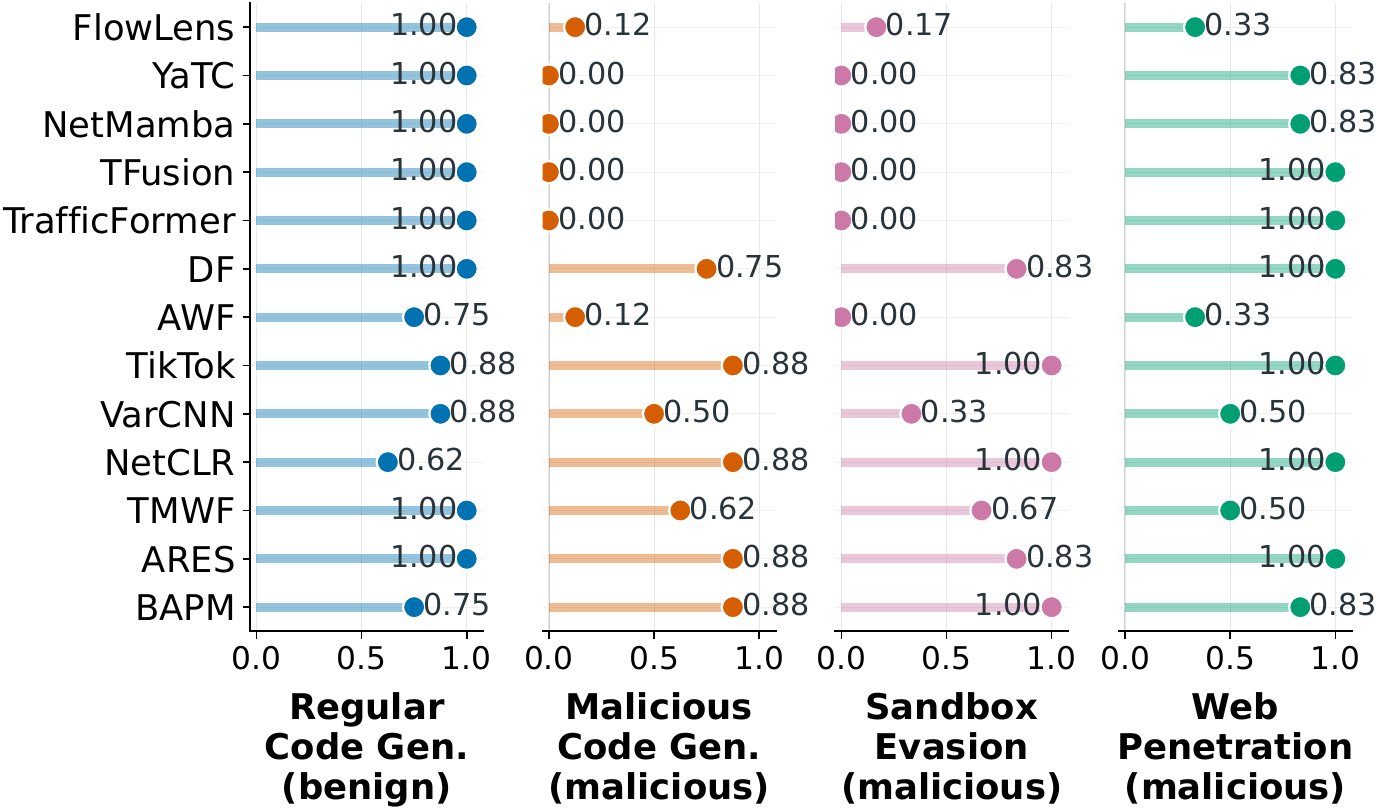}
  \caption{Risk classification by task: specificity for the benign task and recall for malicious tasks.}
  \label{fig:risk-task-recall}
\end{minipage}\hfill
\begin{minipage}[t]{0.48\linewidth}
  \vspace{0pt}
  \centering
  \includegraphics[width=\linewidth,height=1.58in,keepaspectratio]{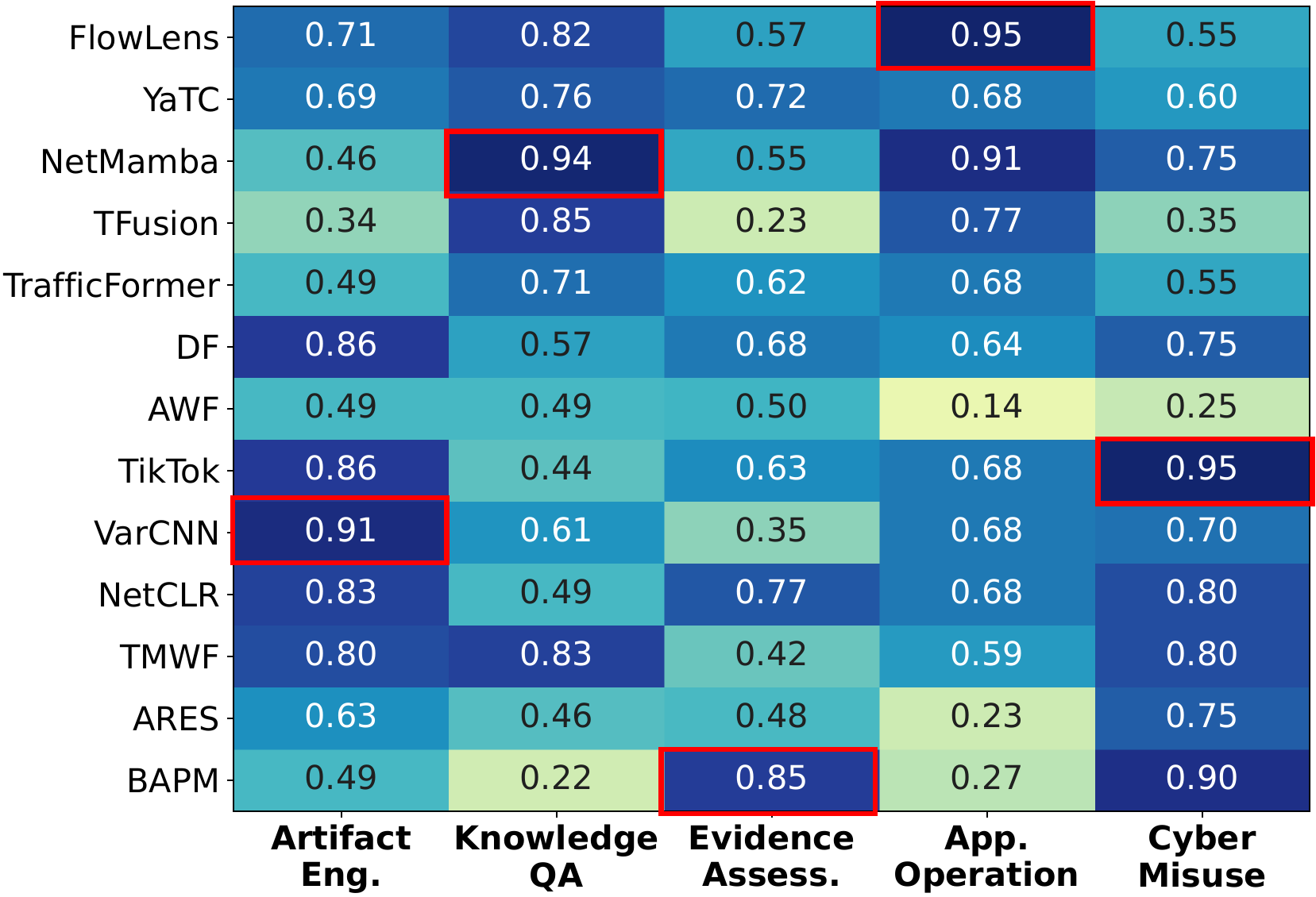}
  \caption{Scenario recognition recall by class.}
  \label{fig:scenario-recall}
\end{minipage}
\par\vspace{1em}
\centering
\newlength{\scenariofigureheight}
\settoheight{\scenariofigureheight}{\includegraphics[width=0.32\linewidth]{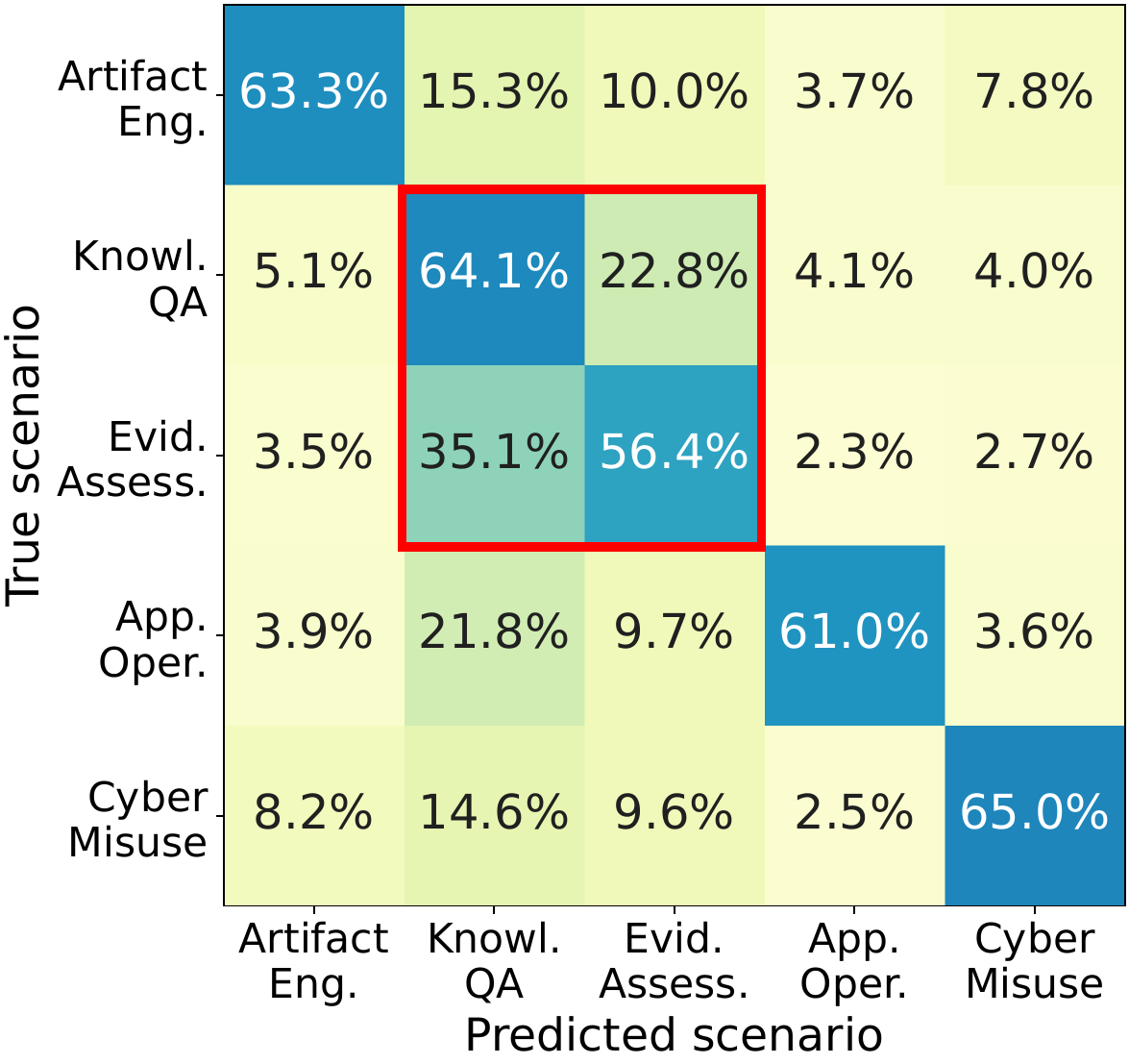}}
\begin{minipage}[t]{0.32\linewidth}
  \vspace{0pt}
  \centering
  \includegraphics[width=\linewidth,height=\scenariofigureheight,keepaspectratio]{sections/figures/scenario_all_baselines_confusion_mod.pdf}
  \captionsetup{width=\linewidth,justification=centering,singlelinecheck=false,skip=\baselineskip}
  \caption{Scenario confusion across all 13 baselines.}
  \label{fig:scenario-confusion}
\end{minipage}\hfill
\begin{minipage}[t]{0.64\linewidth}
  \vspace{0pt}
  \centering
  \includegraphics[width=\linewidth,height=\scenariofigureheight,keepaspectratio]{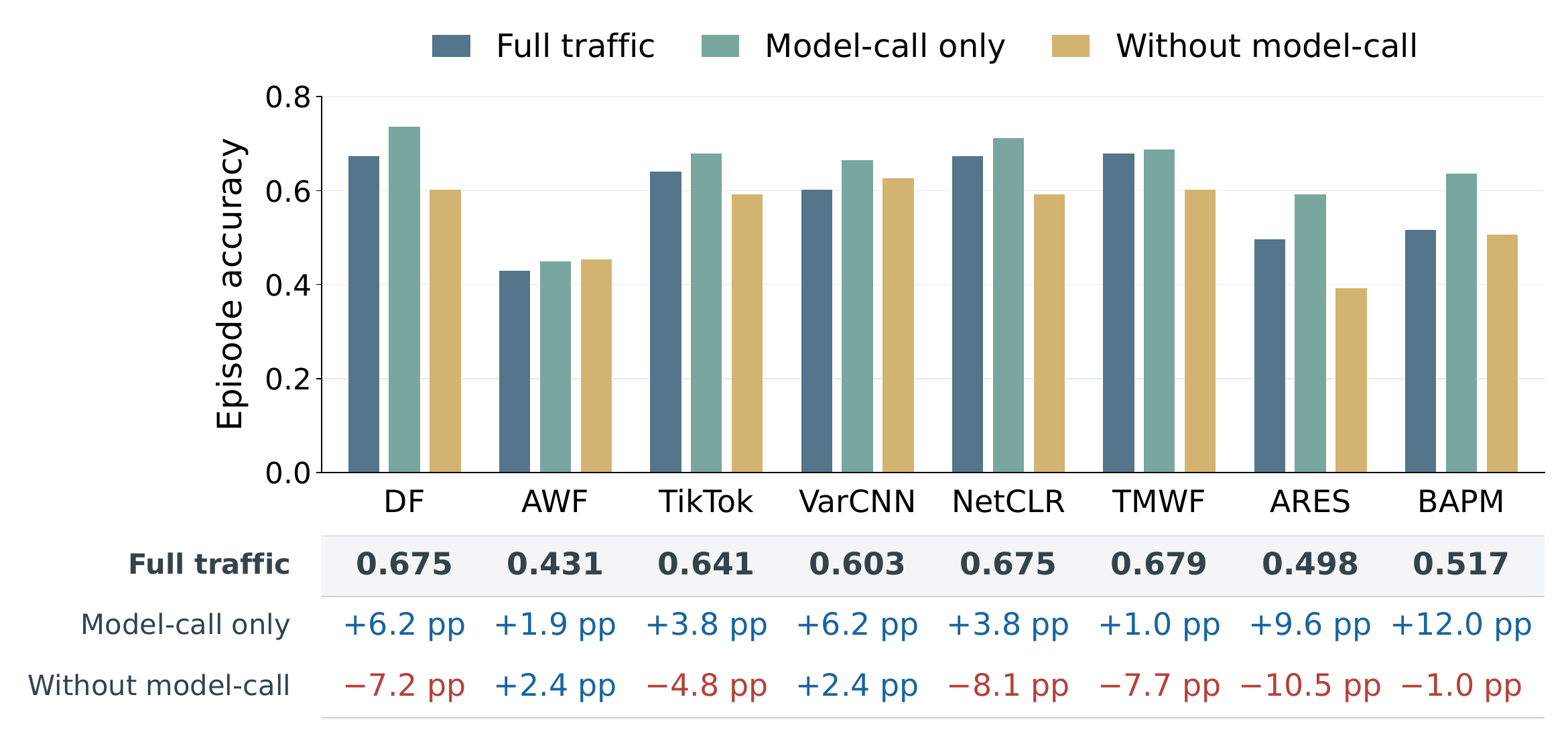}
  \captionsetup{width=\linewidth,justification=centering,singlelinecheck=false,skip=\baselineskip}
  \caption{Scenario recognition accuracy under different traffic scopes.}
  \label{fig:scenario-model-call-ablation}
\end{minipage}
\end{figure}

\begin{figure}[t]
\centering
\vspace{1.2em}
\begin{minipage}[t]{0.35\linewidth}
  \vspace{0pt}
  \centering
  \begin{minipage}[b][1.40in][b]{\linewidth}
    \centering
    \includegraphics[width=\linewidth]{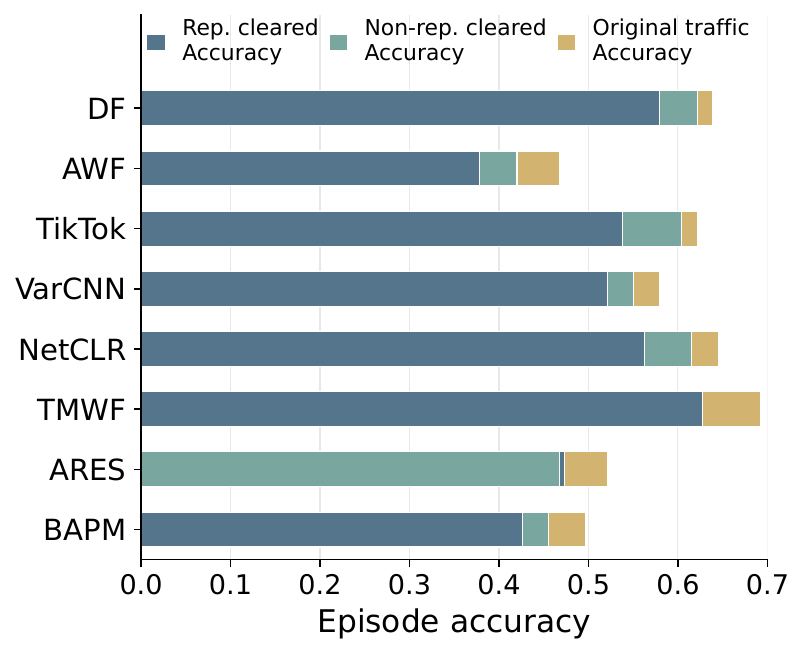}
  \end{minipage}
  \caption{Scenario recognition accuracy after segment clearing.}
  \label{fig:representative-vs-nonrepresentative}
\end{minipage}\hfill
\begin{minipage}[t]{0.63\linewidth}
  \vspace{0pt}
  \centering
  \begin{minipage}[b][1.40in][b]{\linewidth}
    \centering
    \includegraphics[width=\linewidth]{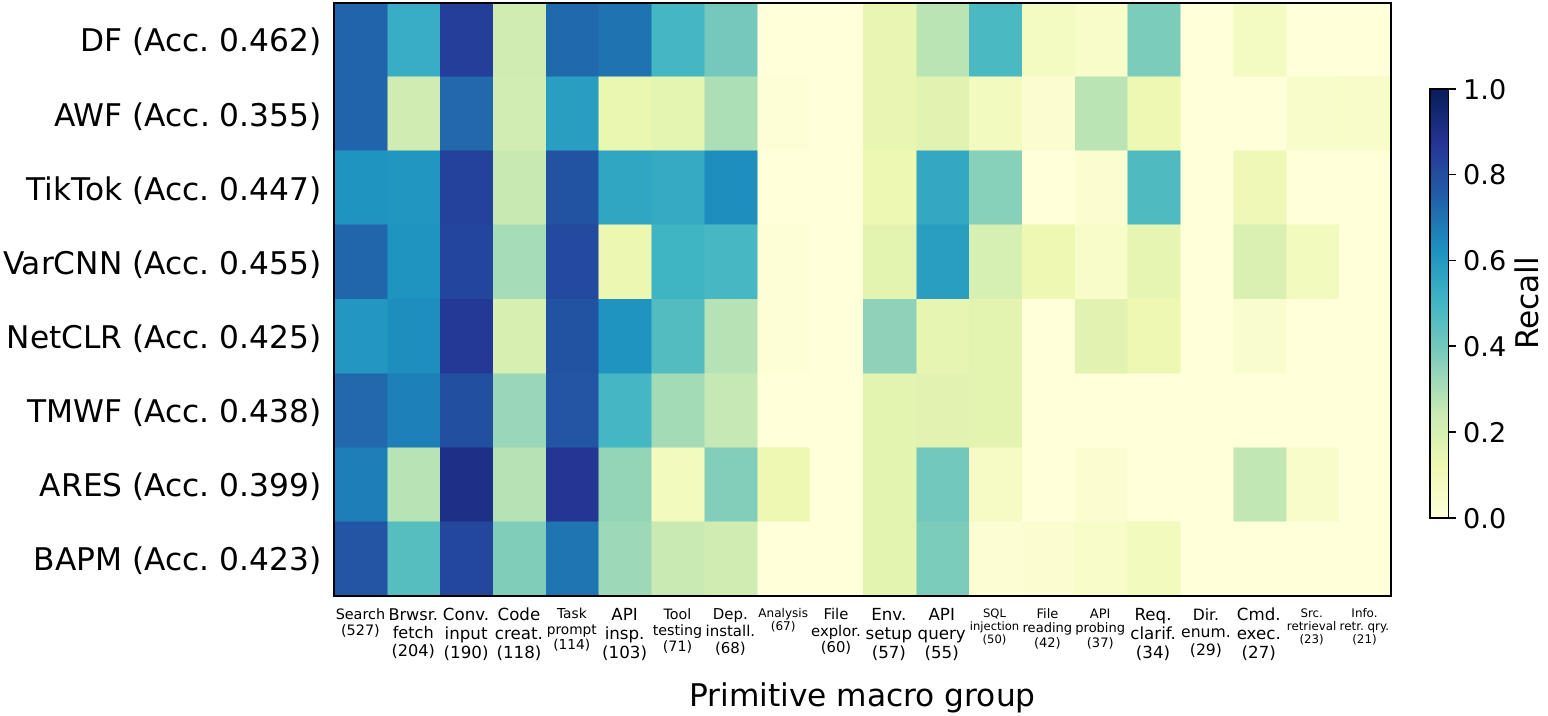}
  \end{minipage}
  \caption{Recall by primitive macro group in behavior primitive classification.}
  \label{fig:primitive-recall}
\end{minipage}
\end{figure}

As shown in Table~\ref{tab:episode-risk-full-traffic}, all 13 baselines achieve accuracy above 90\% on the test set. However, more than half of the baselines achieve recall below 50\%. AWF and FlowLens reach only 15\% and 20\%, respectively.
This gap becomes clearer across individual abusive tasks. Figure~\ref{fig:risk-task-recall} presents recall for three abusive tasks and specificity for the semantically similar regular code generation task. Multiple baselines achieve high recall for web penetration but substantially lower recall for the other two abusive tasks.
Specifically, YaTC, NetMamba, TFusion, and TrafficFormer yield a 0.000 detection rate for malicious code generation and sandbox evasion while achieving a detection rate near 1.000 for web penetration.
In contrast, TikTok, NetCLR, ARES, and BAPM perform relatively well across all three abusive tasks. This result indicates that abusive behaviors leave identifiable signatures in network traffic.

Web penetration's higher recall stems from its direct network interactions like scanning, SQL injection, and XSS leave distinct traffic clues. In contrast, malicious code generation and sandbox evasion rely on code content and execution intent, which encryption obscures. Their observable traffic closely resembles normal coding tasks, differing primarily in repeated security bypass attempts, backend model calls, and strategy adjustments. Identifying these abusive behaviors requires analyzing interaction context, stage sequences, and multi-flow relationships. The detection gap between normal and abusive tasks highlights the difficulty existing models face in recovering this complex risk intent.

\begin{benchmarktakeaway}
\noindent\textbf{Takeaway.} Baseline capability gaps are largest for attacks whose intent is not directly visible in traffic and whose execution resembles benign operations.
\end{benchmarktakeaway}

\subsection{Scenario Recognition Results}
\label{subsec:scenario_results}

Across 13 baselines evaluated on five scenarios, accuracy ranges from 0.431 to 0.727, with NetMamba performing best. No baseline leads recall in all five scenarios (Figure~\ref{fig:scenario-recall}). Encrypted traffic classifiers generally perform better on General Knowledge Question Answering (Knowledge QA) and Interactive Application Operation, while website fingerprinting methods perform better on Computational Artifact Engineering and Cybersecurity Misuse. These complementary strengths show that models recover different subsets of scenario semantics.

Scientific Evidence Assessment and Knowledge QA dominate mutual confusion (Figure~\ref{fig:scenario-confusion}): across pooled baseline predictions, 35.1\% of Evidence Assessment instances are misclassified as Knowledge QA and 22.8\% vice versa.
Both share retrieval, inference, and answer generation, with similar model backend traffic shares (57.44\% and 58.60\%, respectively). Compared with Knowledge QA, Evidence Assessment has 59.5\% longer mean flow duration, 37.4\% more packets, 11.5\% lower mean packet length, and 27.7\% higher inbound payload share. Current baselines do not reliably exploit these differences.

We retrain eight website fingerprinting baselines on full traffic, model calls only, and traffic without model calls (Figure~\ref{fig:scenario-model-call-ablation}). Relative to full traffic evaluated on the same test episodes, model calls alone improve average accuracy by 5.56 percentage points, whereas excluding them reduces it by 4.31 points. The remaining traffic still supports accuracies of 0.392--0.627. Model calls therefore carry strong scenario signals that models capture better in isolation, while other traffic also provides useful information.

To examine how behavior primitives support scenario recognition, we select representative primitives through frequency filtering and maximum cosine similarity between Jina embeddings and the task semantic center, using the closest available primitive when the representative is absent. Clearing one randomly selected representative segment per episode reduces average accuracy by 6.95 percentage points, versus 3.77 points when clearing one randomly selected nonrepresentative segment (Figure~\ref{fig:representative-vs-nonrepresentative}). The larger average decrease after representative segment clearing links task aligned local behavior to scenario predictions.

\begin{benchmarktakeaway}
\noindent\textbf{Takeaway.} Similar scenarios remain difficult to distinguish. Model calls carry strong scenario signals, while task aligned behavior primitive segments provide additional clues for scenario recognition.
\end{benchmarktakeaway}

\subsection{Behavior Primitive Classification Results}
\label{subsec:primitive_macro_group_results}

We evaluate eight website fingerprinting baselines on 2,101 test behavior primitive segments from 47 primitive macro groups.
Overall accuracy is relatively consistent across the baselines, ranging from 0.355 to 0.462. However, performance varies substantially across primitive macro groups. Within a single baseline, recall across groups can range from 0.000 to 0.900.

This imbalance leads to widespread missed detections of long tail primitive macro groups. Figure~\ref{fig:primitive-recall} presents recall for the 20 groups with the most test support.
The ten most frequent groups account for 1,522 of 2,101 behavior primitive segments, or 72.44\%, but contribute between 91.34\% and 96.09\% of all correct predictions made by each model.
Although we upsample rare groups to at least 200 training samples, correct predictions remain concentrated on frequent groups and groups with stable traffic patterns.

Models also struggle to distinguish semantically similar primitive macro groups.
For example, 37 of 204 \texttt{browser\_fetch} segments are predicted as \texttt{search}. Among 527 \texttt{search} segments, 38 are predicted as \texttt{api\_query} and 35 as \texttt{browser\_fetch}.
These errors occur among related retrieval groups. The confusion shows that local traffic patterns carry useful signals, but current baselines struggle to distinguish these primitive macro groups from traffic alone.

\begin{benchmarktakeaway}
\noindent\textbf{Takeaway.} Traffic carries predictive signals for primitive macro groups, but existing baselines classify frequent groups and groups with distinctive traffic patterns more reliably than rare or semantically similar groups.
\end{benchmarktakeaway}

\section{Discussion}
\label{sec:discussion}
\dataset{} supports research on agent behavior inference, transfer learning, and privacy defenses by providing paired traffic and semantic annotations and a common benchmark. We recognize that these auditing capabilities also pose dual-use risks, as attackers could exploit them to infer private user activities. Nevertheless, we believe that systematically studying these capabilities is essential for understanding privacy exposure and developing effective defenses, while enabling responsible oversight of agent use.

Our study has three main limitations. 
First, financial and computational constraints limit coverage of agent applications and configurations, despite including prevalent use cases and diverse tasks. 
Second, AI annotation and grouping can introduce opaque or inaccurate decisions. We mitigate these risks through repeated AI interaction and iterative review and a manual spot check. 
Third, we collect part of \dataset{} in controlled local environments for ethical reasons. However, this may cause traffic timing to differ from that in real deployments. Nevertheless, widely used traffic datasets such as ISCXIDS2012~\citep{shiravi2012benchmark} and CICIDS2017~\citep{sharafaldin2018cicids2017} also rely on controlled testbeds. Their widespread adoption suggests that controlled testbeds can still provide useful benchmarks for comparative model evaluation.

\section{Conclusion}

We construct \dataset{}, a dataset that pairs traffic with episode metadata and behavior primitive annotations, covering 3,114 episodes across 20 tasks and five scenarios, with 276,417 bidirectional flows, 104.07 GB of traffic, and 40,049 primitive segments in 47 macro groups. To our knowledge, it is the first dataset dedicated to network traffic from client-side LLM agent execution.
Our benchmark evaluates 13 methods on risk identification, scenario recognition, and behavior primitive classification. Existing methods capture distinctive traffic patterns but struggle with rare behaviors and semantic distinctions weakly reflected in traffic.
\dataset{} provides a foundation for more precise auditing and forensic analysis of agent behavior from network traffic.

\subsection*{AI Use Statement}
We used AI to assist with dataset annotation, as detailed in Section~\ref{sec:primitive-alignment}. 
We used AI to support code development by delegating routine tasks, such as environment setup and plotting, and use AI to support mathematical calculation, with all outputs reviewed by humans.
We used AI to help identify related work and manually verified the relevance of the cited literature to our work. Finally, we used AI to polish the manuscript, specifically, we wrote the initial draft ourselves and then used AI to refine the language.
The authors take responsibility for the final content of the work.

\subsection*{Ethics Statement}
\noindent\textbf{Data collection.} We designed the collection process to protect privacy and avoid harm to production systems. Tasks were drawn from public benchmarks and involved no real-user accounts, private communications, or personally identifiable information. State-changing benign interactions used locally simulated services, and attack execution was contained in isolated Docker environments. For attack code generation, we retained model safety mechanisms and used only publicly released mutated prompts from Meta's PurpleLlama Cybersecurity Benchmarks, without adding jailbreak, prompt-obfuscation, or safety-bypass strategies. We excluded refusal-only episodes from the task-execution corpus.

\noindent\textbf{Dataset and benchmark release.} We release \dataset{} and its benchmark to support research on agent behavior auditing and privacy protection. The released data contain no private user information. Agent execution logs are sanitized before release to reduce the risk of their reuse for malicious model distillation. These precautions address the potential misuse of execution traces alongside the research benefits of public access. We release the code we develop under the GNU General Public License version 3 (GPLv3), as specified in \texttt{LICENSE}. We release our contributions to the \dataset{} data, including network traffic traces, sanitized agent execution logs, metadata, and annotations, under the Creative Commons Attribution 4.0 International license (CC BY 4.0), as specified in \texttt{LICENSE-DATA}. This data license does not cover software code or materials from third parties. Source benchmark prompts and dependencies remain subject to their respective licenses.

\subsection*{Reproducibility Statement}
Section~\ref{sec:ant-design} describes the data collection and annotation pipeline, while Section~\ref{sec:benchmark_evaluation} presents the benchmark objectives, baseline methods, evaluation metrics, and runtime environment. Appendix~\ref{app:experiment-configuration} provides detailed data split statistics, training hyperparameters, and prediction aggregation strategies. 

We release the packet-level ANT dataset, data collection and benchmark suites at \url{https://anonymous.4open.science/r/ant-main-suite-7BC0/}. These resources support independent evaluation of the reported baselines and further research on agent behavior auditing from network traffic.

\clearpage
\appendix
\section{Task Composition and Collection Configuration}
\label{app:scenario-task-composition}

\subsection{Scenario and Task Composition}
\label{app:task-composition}

\begin{center}
    \centering
    \fontsize{8}{9}\selectfont
    \captionof{table}{The 20 tasks in \dataset{}, organized into the five scenarios used for scenario recognition. References identify the source paper and versioned dataset or repository where available.}
    \label{tab:flowbenchmark_task_scenarios}
    \setlength{\tabcolsep}{4pt}
    \renewcommand{\arraystretch}{1.06}
    \begin{tabularx}{\textwidth}{@{}>{\raggedright\arraybackslash}p{0.26\textwidth}|>{\raggedright\arraybackslash}X@{}}
    \toprule
    \textbf{Scenario} & \textbf{Task} \\
    \midrule

    \textbf{Computational Artifact Engineering:}
    agents inspect, modify, generate, or validate code and data artifacts.
    & \textbf{SWE-bench Lite bug fixing:} the agent checks out a real repository, reads an issue, edits code, runs tests, and produces a patch~\citep{jimenez2024swebench,princetonnlp2023swebenchlite}.\par
      \textbf{SWE-bench Verified code review:} we convert human-validated SWE-bench issue and patch pairs into review tasks in which the agent inspects the patch, optionally runs tests, and returns review comments~\citep{jimenez2024swebench,princetonnlp2024swebenchverified}.\par
      \textbf{DS-1000 data science completion:} the agent fills the missing code at an \texttt{[insert]} placeholder in a data-science program~\citep{lai2023ds1000}.\par
      \textbf{DA-Dev data analysis:} the agent performs an end-to-end Python data analysis workflow, including reading data, processing it, modeling, and reporting results~\citep{hu2024infiagent}. \\
    \midrule

    \textbf{General Knowledge Question Answering:}
    agents answer general knowledge questions through factual retrieval, multi-step reasoning, and answer synthesis.
    & \textbf{GAIA general assistant:} the agent solves cross-step reasoning problems using search, local files, Python, and answer synthesis. We use the official 2023 GAIA release at the recorded repository revision~\citep{mialon2024gaia,gaiabenchmark2023dataset}.\par
      \textbf{Natural Questions web QA:} the agent searches the web, opens relevant pages, and extracts answers to factual questions~\citep{kwiatkowski2019naturalquestions}.\par
      \textbf{GAIA filtered text-only web QA:} the agent answers GAIA questions that have no attached file. The source is \texttt{sayan1101/gaia\_filtered\_text\_only}, revision \texttt{8b387ca}, with 127 validation and 230 test records~\citep{mialon2024gaia,sayan2023gaiatextonly}.\par
      \textbf{FRAMES multi-step factual reasoning:} the agent performs chained web searches to combine facts across steps and derive the answer~\citep{krishna2024frames}.\par
      \textbf{HotpotQA multi-hop QA:} the agent retrieves evidence from multiple sources and combines it to answer multi-hop questions~\citep{yang2018hotpotqa}.\par
      \textbf{ASQA long-answer QA:} the agent resolves ambiguous questions through multi-branch search and writes a cited long-form answer~\citep{stelmakh2022asqa}. \\
    \midrule

    \textbf{Scientific Evidence Assessment:}
    agents retrieve, verify, and synthesize scientific evidence to answer scholarly questions or assess claims.
    & \textbf{QASPER paper QA:} the agent searches for a target paper, reads its abstract or full text, and answers paper-specific questions~\citep{dasigi2021qasper}.\par
      \textbf{SciFact scientific fact checking:} the agent searches for evidence papers, extracts supporting or refuting evidence, and verifies a scientific claim~\citep{wadden2020scifact}.\par
      \textbf{PubMedQA biomedical QA:} the agent searches biomedical literature and decides whether the answer is yes, no, or maybe~\citep{jin2019pubmedqa}.\par
      \textbf{ScholarQA literature synthesis:} the agent searches across multiple papers and synthesizes evidence into a literature-style answer~\citep{asai2024openscholar}.\par
      \textbf{ORKG SciQA scholarly KG QA:} the agent answers academic questions through search-backed retrieval and synthesis~\citep{auer2023sciqa}. \\
    \midrule

    \textbf{Interactive Application Operation:}
    agents fulfill operational requests through stateful interactions with applications and service APIs.
    & \textbf{WebArena web navigation:} the agent uses a browser to complete tasks such as repository operations, forum posting, and shopping in web environments~\citep{zhou2024webarena}.\par
      \textbf{MultiWOZ office assistant:} we adapt MultiWOZ 2.2 dialogue goals for attractions and train booking to a local assistant environment with contact, calendar, and email APIs~\citep{zang2020multiwoz}. \\
    \midrule

    \textbf{Cybersecurity Misuse:}
    agents use code generation and execution capabilities to produce offensive behavior, bypass restrictions, or exploit vulnerabilities.
    & \textbf{MITRE ATT\&CK code generation:} the agent writes code intended to implement attack techniques from the MITRE ATT\&CK framework~\citep{bhatt2023purplellama,bhatt2024cyberseceval2}.\par
      \textbf{Interpreter abuse:} the agent attempts to execute Python in a restricted sandbox and bypass filters or sandbox boundaries~\citep{bhatt2024cyberseceval2}.\par
      \textbf{OWASP Juice Shop penetration testing:} the agent uses a browser and HTTP tools against an isolated vulnerable web application to complete web-security challenges~\citep{owaspjuiceshop}. \\
    \bottomrule
    \end{tabularx}
\end{center}

\clearpage

\subsection{Agent Configurations}
\label{app:agent-configurations}

Table~\ref{tab:agent-configurations} specifies the agent software and backend model for each of the five configurations used in \dataset{}. 

\begin{center}
    \small
    \captionof{table}{Agent software versions and backend models used for data collection.}
    \label{tab:agent-configurations}
    \setlength{\tabcolsep}{5pt}
    \renewcommand{\arraystretch}{1.15}
    \begin{tabularx}{\textwidth}{@{}l l l >{\raggedright\arraybackslash}X@{}}
    \toprule
    Configuration & Agent software & Version & Backend model \\
    \midrule
    \texttt{cc\_dpsk} & \href{https://github.com/anthropics/claude-code/releases/tag/v2.1.212}{Claude Code} & 2.1.212 & \href{https://api-docs.deepseek.com/updates/}{DeepSeek-V4-Flash} \\
    \texttt{cc\_mimo} & Claude Code & 2.1.212 & \href{https://mimo.xiaomi.com/mimo-v2-pro}{MiMo-V2-Pro} \\
    \texttt{cc\_sonnet} & Claude Code & 2.1.212 & \href{https://www.anthropic.com/claude/sonnet}{Claude Sonnet 4.6} \\
    \texttt{codex} & \href{https://github.com/openai/codex/releases/tag/rust-v0.135.0}{Codex CLI} & 0.135.0 & \href{https://developers.openai.com/api/docs/models/gpt-5.4}{GPT-5.4} (medium reasoning) \\
    \texttt{qwen} & \href{https://github.com/QwenLM/qwen-code/releases/tag/v0.17.1}{Qwen Code} & 0.17.1 & \href{https://docs.modelstudio.console.alibabacloud.com/zh/model-studio/qwen3-6-plus}{Qwen3.6-Plus} \\
    \bottomrule
    \end{tabularx}
\end{center}

\subsection{Collection Server Environments}
\label{app:collection-servers}

We collect traffic on two Alibaba Cloud servers. Table~\ref{tab:collection-server-environments} summarizes their hardware and operating systems.

\begin{center}
    \small
    \captionof{table}{Hardware and software environments of the data collection servers.}
    \label{tab:collection-server-environments}
    \setlength{\tabcolsep}{5pt}
    \renewcommand{\arraystretch}{1.18}
    \begin{tabularx}{\textwidth}{@{}l >{\raggedright\arraybackslash}X >{\raggedright\arraybackslash}X@{}}
    \toprule
    Component & Domestic server & Overseas server \\
    \midrule
    Processor & Intel Xeon 6982P-C & Intel Xeon 6982P-C \\
    CPU allocation / Memory & 8 vCPUs / 32\,GB & 8 vCPUs / 16\,GiB \\
    Operating system & Ubuntu 22.04 & Alibaba Cloud Linux 4.0.3 (OpenAnolis Edition) \\
    Kernel & 5.15.0-177-generic & 6.6.102-5.3.3.\allowbreak alnx4.x86\_64 \\
    \bottomrule
    \end{tabularx}
\end{center}

\clearpage
\section{Behavior Primitive Annotation Details}
\label{app:primitive-details}

\subsection{Semantic Annotation Prompt}
\label{app:primitive-annotation-prompt}
The following prompt is used for semantic annotation with DeepSeek-v4-pro.

\begin{Verbatim}[fontsize=\footnotesize,breaklines=true,breakanywhere=true,breaksymbolleft={}]
You are annotating grouped agent-session logs for a research dataset.

Your job is semantic labeling. The script has already grouped message fragments,
merged assistant chunks, and attached tool results to the originating message.
Do not derive labels by keyword matching. Read the message content, tool calls,
and tool results, then summarize the message's high-level intent in English.

Intent granularity examples:
- API endpoint probing or documentation discovery
- registering/logging in a test account
- extracting current user id, basket id, token, or cookie state
- re-querying an API after new evidence
- checking challenge/status output
- browser page visit or browser network capture
- checking response content, file content, or command output
- answer synthesis, report writing, or final result formatting

Label each message independently while using surrounding task context. For very
short operational messages, summarize what that message accomplishes in the task
flow. If the evidence is ambiguous, say so briefly instead of inventing details.

Return exactly one annotation per input message. Each annotation must contain `message_index`, `message_id`, `role`, `intent`, `attack_primitive`, `primitive_zh`, and `evidence`. Use attack primitives as retrospective labels of observed behavior. For non-security tasks under this forced field name, use neutral workflow primitives such as paper_search, data_analysis, code_debugging, answer_synthesis, or browser_navigation. Keep `intent`, `attack_primitive`, `primitive_zh`, and `evidence` concise: one short phrase or sentence each. Do not include raw long responses, code blocks, or URLs unless essential.

Output must be valid JSON with this top-level shape:
{"annotations":[{...}, ...]}
Do not include markdown, comments, or extra prose.
\end{Verbatim}

\clearpage
\subsection{Primitive Macro Groups}
\label{app:primitive-macro-groups}

Table~\ref{tab:primitive-macro-groups} lists the canonical labels
and split counts for the 47 macro groups. Counts include all
macro-group-annotated segments, including 11 empty segments
and 8 nonempty segments without usable TCP packets, but exclude
synthetic prelude segments. No training oversampling is applied
to these counts.

\begingroup
\footnotesize
\setlength{\tabcolsep}{3pt}
\begin{longtable}{llrrrr}
\caption{Primitive macro groups and split counts, including empty
segments and segments without usable TCP packets, excluding
synthetic prelude segments. Group IDs refer to the released
primitive-to-group mapping.}
\label{tab:primitive-macro-groups}\\
\toprule
Group & Canonical label & Primitive types & Train & Validation & Test \\
\midrule
\endfirsthead
\toprule
Group & Canonical label & Primitive types & Train & Validation & Test \\
\midrule
\endhead
N001 & analysis & 90 & 1308 & 152 & 67 \\
N002 & api\_discovery & 73 & 531 & 34 & 19 \\
N003 & api\_inspection & 91 & 1408 & 130 & 103 \\
N004 & api\_probing & 91 & 671 & 21 & 37 \\
N005 & api\_query & 95 & 1685 & 73 & 55 \\
N006 & application\_reconnaissance & 86 & 451 & 26 & 8 \\
N007 & authentication\_token\_acquisition & 86 & 300 & 2 & 3 \\
N008 & browser\_fetch & 71 & 3042 & 225 & 204 \\
N009 & bypass\_attempt & 71 & 170 & 3 & 2 \\
N010 & challenge\_info\_query & 61 & 279 & 27 & 20 \\
N011 & code\_creation & 107 & 2930 & 156 & 118 \\
N012 & command\_execution & 37 & 759 & 45 & 27 \\
N013 & configuration\_manipulation & 62 & 104 & 3 & 4 \\
N014 & conversation\_input & 107 & 2400 & 173 & 191 \\
N015 & credential\_attempt & 140 & 419 & 3 & 10 \\
N016 & data\_decoding & 17 & 76 & 12 & 2 \\
N017 & data\_extraction & 62 & 257 & 27 & 16 \\
N018 & dependency\_check & 33 & 203 & 22 & 4 \\
N019 & dependency\_installation & 42 & 724 & 45 & 68 \\
N020 & directory\_listing\_enumeration & 76 & 354 & 16 & 29 \\
N021 & environment\_setup & 63 & 1123 & 106 & 57 \\
N022 & error\_diagnosis & 44 & 71 & 11 & 13 \\
N023 & exploit\_attempt & 66 & 444 & 1 & 5 \\
N024 & file\_download & 32 & 114 & 7 & 3 \\
N025 & file\_exploration & 109 & 668 & 24 & 60 \\
N026 & file\_reading & 56 & 603 & 53 & 42 \\
N027 & http\_header\_probe & 14 & 28 & 2 & 1 \\
N028 & information\_retrieval\_query & 82 & 703 & 33 & 21 \\
N029 & mock\_server\_alternative\_start & 46 & 71 & 10 & 0 \\
N030 & network\_check & 45 & 176 & 59 & 6 \\
N031 & payload\_crafting & 142 & 253 & 20 & 8 \\
N032 & planning & 69 & 193 & 16 & 4 \\
N033 & repository\_checkout & 76 & 309 & 54 & 12 \\
N034 & requirement\_clarification & 47 & 333 & 9 & 34 \\
N035 & sandbox\_probing & 66 & 160 & 2 & 8 \\
N036 & script\_writing & 33 & 259 & 15 & 1 \\
N037 & search & 104 & 6169 & 578 & 527 \\
N038 & source\_code\_retrieval & 58 & 1094 & 10 & 23 \\
N039 & sql\_injection & 70 & 403 & 43 & 50 \\
N040 & status\_check & 98 & 229 & 16 & 6 \\
N041 & task\_prompt & 96 & 2011 & 144 & 114 \\
N042 & tool\_testing & 110 & 1173 & 87 & 71 \\
N043 & user\_booking\_request & 28 & 50 & 7 & 4 \\
N044 & verification & 113 & 307 & 14 & 8 \\
N045 & version\_control\_diff\_check & 47 & 151 & 6 & 15 \\
N046 & workflow\_instructions & 50 & 139 & 6 & 20 \\
N047 & xss\_testing & 33 & 113 & 1 & 2 \\
\midrule
Total & & 3295 & 35418 & 2529 & 2102 \\
\bottomrule
\end{longtable}
\endgroup

\begin{figure}[htbp]
    \centering
    \includegraphics[width=\textwidth]{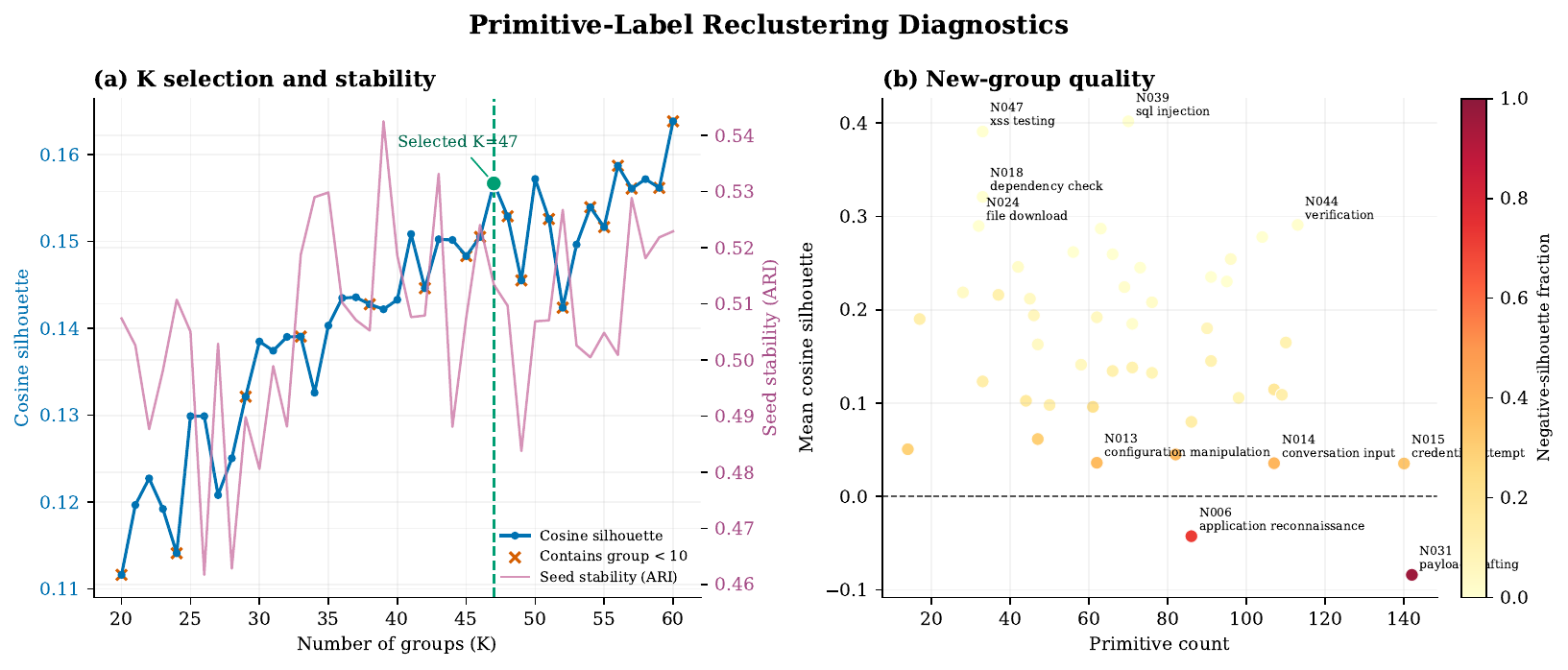}
    \caption{Primitive clustering diagnostics. Left: cosine silhouette and seed stability across candidate $K$, with crosses marking partitions containing groups of fewer than 10 primitive types. Right: macro group size, mean cosine silhouette, and the fraction of primitive types with negative silhouette at $K=47$. Group IDs match Table~\ref{tab:primitive-macro-groups}.}
    \label{fig:primitive-reclustering-diagnostics}
\end{figure}

\clearpage
Figure~\ref{fig:primitive-reclustering-diagnostics} shows the clustering diagnostics. At $K=47$, the cosine silhouette is approximately 0.157 and seed stability is approximately 0.51 ARI; the smallest macro group contains 14 primitive types.

\section{Scenario Cohesion of Primitive Distributions and Sequences}
\label{app:ant-primitive-structure}

We analyze scenario-related primitive structure from two complementary perspectives, i.e., distributional cohesion and sequence organization beyond composition.
For an episode $i$ of length $n_i$, let
$z_{it}\in\{1,\ldots,K\}$ denote the primitive macro group at position $t$,
and let $x_{it}\in\mathbb R^{1024}$ denote its unit-normalized primitive
embedding. We use the current grouping with $K=47$ and fixed Jina embeddings
of the original primitive labels. 

Comparisons first average episode distances within each pair of distinct
tasks, then give equal weight to scenarios and settings. Let
$\mathcal E_{a,t}$ contain the episodes of task $t$ under setting $a$, and let
$\mathcal T_s$ contain the tasks in scenario $s$. Define
\begin{equation}
 \bar d_a(t,u)=\frac{1}{|\mathcal E_{a,t}|\,|\mathcal E_{a,u}|}
    \sum_{i\in\mathcal E_{a,t}}\sum_{j\in\mathcal E_{a,u}}d(i,j),
 \qquad t\ne u.
 \label{eq:ant-task-distance}
\end{equation}
For $M$ eligible settings and $S=5$ scenarios, aggregate distances are
\begin{align}
 D_{\mathrm{within}}
 &=\frac{1}{MS}\sum_a\sum_s
   \frac{\sum_{\{t,u\}\subset\mathcal T_s}\bar d_a(t,u)}
        {\binom{|\mathcal T_s|}{2}},
 \label{eq:ant-within}\\
 D_{\mathrm{between}}
 &=\frac{1}{M\binom{S}{2}}\sum_a\sum_{s<r}
   \frac{\sum_{t\in\mathcal T_s}\sum_{u\in\mathcal T_r}\bar d_a(t,u)}
        {|\mathcal T_s|\,|\mathcal T_r|}.
 \label{eq:ant-between}
\end{align}
The within-scenario sum uses unordered pairs of distinct tasks. We report
$G_d=1-D_{\mathrm{within}}/D_{\mathrm{between}}$ after aggregating distances
across settings. Positive gains quantify how much closer different tasks
are within the same scenario. All reported estimates use this aggregate
comparison.

\subsection{Primitive Distribution Cohesion}
\label{app:ant-primitive-distributions}

\paragraph{Distribution geometry.}
We compare primitive distributions in both continuous semantic space and
discrete category space. The empirical measure
$\mu_i=n_i^{-1}\sum_{t=1}^{n_i}\delta_{x_{it}}$ assigns probability mass to each
primitive embedding according to its occurrence count. We use the MMD
framework of \citet{ant_gretton2012kernel} to compare these measures with
Equation~\ref{eq:ant-primitive-mmd}. The kernel is
\begin{equation}
 k(x,y)=\frac13\sum_{c\in\{1/2,1,2\}}
           \exp\!\left(-\frac{\|x-y\|_2^2}{2c^2\sigma^2}\right).
 \label{eq:ant-kernel}
\end{equation}
This characteristic kernel distinguishes probability distributions through
their kernel means. Its bandwidth is chosen independently of scenario labels.
We sample 200,000 pairs of primitive vectors from the annotated corpus
and use the median nonzero squared distance, $\sigma^2=0.9569$.
This corpus-level bandwidth is fixed for the analysis cohort, which contains
2,564 distinct primitive labels, a subset of the 3,295 distinct primitive
labels in the full annotated corpus (Section~\ref{sec:primitive-alignment}).
Expanding the squared norm gives
\begin{align}
 d_k^2(\mu_i,\mu_j)
 &=\frac{1}{n_i^2}\sum_{r,t=1}^{n_i} k(x_{ir},x_{it})
   +\frac{1}{n_j^2}\sum_{r,t=1}^{n_j} k(x_{jr},x_{jt})\nonumber\\
 &\quad-\frac{2}{n_i n_j}\sum_{r=1}^{n_i}\sum_{t=1}^{n_j}
                  k(x_{ir},x_{jt}).
 \label{eq:ant-empirical-mmd}
\end{align}
We retain diagonal terms and take the nonnegative square root to obtain the
exact distance between the empirical measures. MMD yields a cohesion gain
of 15.46\%, indicating that different tasks within a scenario have more
similar primitive embedding distributions.

We examine the same distributional relationship through optimal transport.
Sliced-Wasserstein distance averages one-dimensional transport costs over
random projections \citep{ant_nadjahi2020sliced}. Draw $L$ unit directions
$\theta_\ell$ uniformly from the sphere and hold them fixed. Let
$F_{i,\theta_\ell}^{-1}$ denote the quantile function of
$\theta_\ell^\top x$ under $\mu_i$. Then
\begin{equation}
 \operatorname{SW}_{2,L}^2(\mu_i,\mu_j)
 =\frac1L\sum_{\ell=1}^{L}\int_0^1
   \left|F_{i,\theta_\ell}^{-1}(u)-F_{j,\theta_\ell}^{-1}(u)\right|^2du.
 \label{eq:ant-sliced}
\end{equation}
Each integral is evaluated exactly over the union of empirical quantile
breakpoints. The finite direction average approximates the spherical
integral. With 512 directions, the cohesion gain is 11.26\%. Increasing the
direction budget to 1,024 gives 11.17\%, and using 256 gives 11.62\%.
The direction sets share a fixed prefix, and the resulting gains differ by
0.447 percentage points across budgets. This transport comparison corroborates
the semantic distributional cohesion observed with MMD.

To connect continuous semantics with the primitive macro group representation,
let $p_i(z)=n_{iz}/n_i$ be the proportion of group $z$ in episode $i$. We use
Hellinger distance,
\begin{equation}
 H^2(p_i,p_j)=\frac12\sum_{z=1}^{K}
       \left(\sqrt{p_i(z)}-\sqrt{p_j(z)}\right)^2
       =1-\sum_{z=1}^{K}\sqrt{p_i(z)p_j(z)}.
 \label{eq:ant-hellinger}
\end{equation}
The square-root transformation maps the probability simplex to the positive
orthant of a unit sphere. Hellinger distance is the corresponding scaled
chord distance and directly accommodates zero frequencies. Using $H$ in the
cohesion statistic gives a 14.79\% gain. The MMD and Hellinger associations
both pass the corrected scenario-label tests in Table~\ref{tab:ant-cohesion}.
Together with the transport result, they show that scenario cohesion is
expressed in both primitive semantics and category proportions.

For all four primary representations, we assess scenario association by
randomly reassigning scenario labels to tasks 9,999 times. Each reassignment
preserves the original five scenario task counts and assigns the
same label to each task across settings. 
The permutation $p$ value compares
the observed gain with those obtained from these alternative task groupings,
using one plus the exceedance count divided by one plus the permutation count.
We estimate execution variability by resampling episodes within each setting
and task 1,999 times. Each resample updates all pairwise distances through
shared episode weights. Its 2.5th and 97.5th gain percentiles define the
95\% interval for the observed task inventory. The four scenario-association
tests and the two additional-order tests in
Section~\ref{app:ant-primitive-sequences} are adjusted together using the Holm
procedure. Table~\ref{tab:ant-cohesion} reports the aggregate estimates,
intervals, and adjusted $p$ values.

\begin{table}[t]
\centering
\small
\caption{Aggregate scenario cohesion across distinct tasks. Intervals summarize
execution variability within the observed task inventory. The adjusted
$p_{\mathrm H}$ values account for the six primary tests. Distances use the
scale of each representation, and gains report relative cohesion.}
\label{tab:ant-cohesion}
\begin{tabular}{@{}lrrrr@{}}
\toprule
Representation & $D_{\mathrm{within}}$ & $D_{\mathrm{between}}$
 & $G_d$ [95\% CI] (\%) & $p_{\mathrm H}$ \\
\midrule
Primitive embedding MMD & 0.431680 & 0.510646 & 15.46 [14.32, 16.64] & 0.0006 \\
Primitive macro group Hellinger & 0.698573 & 0.819826 & 14.79 [13.98, 15.59] & 0.0050 \\
Primitive embedding area & 0.147906 & 0.158600 & 6.74 [5.85, 7.68] & 0.0182 \\
Primitive macro group residual & 0.389167 & 0.412703 & 5.70 [5.09, 6.36] & 0.0182 \\
\bottomrule
\end{tabular}
\end{table}

\paragraph{Scenario information in composition.}
The distance comparisons establish aggregate cohesion. An entropy decomposition
then quantifies the composition information associated with scenario.
Sample setting $B$, scenario $Y$, task $T$, and episode $E$ with the balanced
weights, followed by a uniformly selected primitive occurrence within the
episode. Denote its primitive macro group by $Z$. The nesting of episodes
within tasks and tasks within scenarios gives
\begin{equation}
 I(Z;E\mid B)
 =I(Z;Y\mid B)+I(Z;T\mid Y,B)+I(Z;E\mid T,B).
 \label{eq:ant-entropy-chain}
\end{equation}
To derive the identity, insert and subtract $H(Z\mid Y,B)$ and
$H(Z\mid T,B)$ between $H(Z\mid B)$ and $H(Z\mid E,B)$.
The three differences are nonnegative conditional mutual information terms.
The left-hand side is the weighted Jensen--Shannon variation of the empirical
episode composition distributions, measured in bits.

\begin{table}[t]
\centering
\small
\caption{Aggregate decomposition of empirical primitive macro group composition.
Components are averaged across settings before dividing by the average total.}
\label{tab:ant-js}
\begin{tabular}{@{}lrr@{}}
\toprule
Component & Information in bits & Share of total in percent \\
\midrule
Scenario & 0.7703 & 34.86 \\
Task within scenario & 0.5045 & 22.84 \\
Episode within task & 0.9346 & 42.30 \\
\midrule
Total & 2.2094 & 100.00 \\
\bottomrule
\end{tabular}
\end{table}

The scenario component is 0.7703 bits, accounting for 34.86\% of total
empirical composition variation. Table~\ref{tab:ant-js} gives the complete
additive decomposition. This quantifies the scenario-associated component
of primitive usage alongside the distance reductions above.

\subsection{Primitive Sequence Structure}
\label{app:ant-primitive-sequences}

\paragraph{Composition-conditioned order representations.}
We next ask whether primitive organization carries scenario structure beyond
its composition. The continuous and discrete representations admit complementary
order statistics with exact expectations under random permutation.
For primitive embeddings, we use a second-level path signature
\citep{ant_chevyrev2016signature}. Draw a fixed
$R\in\mathbb R^{32\times1024}$ with independent $\mathcal N(0,1/32)$ entries,
and construct a piecewise linear cumulative path from the origin using
increments $v_{it}=(1,Rx_{it})/n_i$. The first coordinate records normalized
sequence progress. Its first two signature levels satisfy
\begin{equation}
 S_i^{(1)}=\sum_t v_{it},\qquad
 S_i^{(2)}=\sum_{r<t}v_{ir}\otimes v_{it}
                      +\frac12\sum_t v_{it}\otimes v_{it}.
 \label{eq:ant-signature}
\end{equation}
Expanding $S_i^{(1)}\otimes S_i^{(1)}$ yields
\begin{equation}
 \operatorname{Sym}(S_i^{(2)})
 =\frac12 S_i^{(1)}\otimes S_i^{(1)},\qquad
 A_i=\frac12\bigl(S_i^{(2)}-(S_i^{(2)})^\top\bigr).
 \label{eq:ant-signature-split}
\end{equation}
Fixing composition fixes the endpoint and therefore the symmetric part.
The antisymmetric component retains directional ordering information,
\begin{equation}
 A_i=\frac12\sum_{r<t}
       (v_{ir}\otimes v_{it}-v_{it}\otimes v_{ir}),
 \qquad
 \mathbb E_{\pi}[A_i^{\pi}\mid\{v_{it}\}_{t=1}^{n_i}]=0.
 \label{eq:ant-primitive-area}
\end{equation}
Under a uniform permutation $\pi$, either member of an indexed pair of
increments occurs first with probability $1/2$, so the antisymmetric
contributions cancel in expectation. Progress and semantic coordinates capture
early or late semantic content, while pairs of semantic coordinates capture
their relative ordering. We compare these second-level order summaries using
$\|A_i-A_j\|_F$.

For primitive macro groups, we apply the same composition-conditioned principle
to adjacent transitions. Let
\begin{equation}
 J_i(a,b)=\frac{1}{n_i-1}\sum_{t=1}^{n_i-1}
             \mathbf1\{z_{it}=a,\ z_{i,t+1}=b\}.
 \label{eq:ant-transition}
\end{equation}
Given fixed group counts, uniform permutation places group $a$ at the first
position of a pair with probability $n_{ia}/n_i$. The conditional probability
of group $b$ at its second position is
$(n_{ib}-\mathbf1\{a=b\})/(n_i-1)$. Subtracting this exact expectation gives
\begin{equation}
 C_i(a,b)=J_i(a,b)
       -\frac{n_{ia}(n_{ib}-\mathbf1\{a=b\})}{n_i(n_i-1)},
 \qquad
 \mathbb E_{\pi}[C_i^{\pi}\mid\{n_{ia}\}]=0.
 \label{eq:ant-residual}
\end{equation}
We compare episodes using $\|C_i-C_j\|_F$. The residual's symmetric component
summarizes adjacency preferences, and its antisymmetric component summarizes
directional preferences. Their orthogonality gives
\begin{equation}
 C_i^+=\frac{C_i+C_i^\top}{2},\qquad
 C_i^-=\frac{C_i-C_i^\top}{2},\qquad
 \|C_i\|_F^2=\|C_i^+\|_F^2+\|C_i^-\|_F^2.
 \label{eq:ant-transition-orthogonal}
\end{equation}
Thus directed area describes semantic ordering, and transition residuals
describe the organization of primitive macro groups. Their aggregate gains
are 6.74\% and 5.70\%, respectively, with Holm-adjusted $p=0.0182$ for both
scenario-association tests.

\paragraph{Evidence from order randomization.}
We evaluate the additional contribution of observed order by comparing these
gains with a reference distribution that preserves every episode's primitive
multiset and length. In each of 999 trials, we independently permute positions
within each episode and recompute the sequence representations, distances,
and aggregate gain. This preserves composition-dependent variability in the
randomized comparison. Define the additional gain and its permutation
$p$ value as
\begin{equation}
 \Delta G_d=G_d^{\mathrm{obs}}-\frac1P\sum_{b=1}^{P}G_d^{(b)},
 \qquad
 \widehat p=\frac{1+\sum_{b=1}^{P}\mathbf1\{G_d^{(b)}\geq G_d^{\mathrm{obs}}\}}
                 {P+1},
 \quad P=999.
 \label{eq:ant-order-test}
\end{equation}
Directed area exceeds its randomized reference by 5.77 percentage points,
and transition residuals exceed theirs by 5.24 percentage points.
Both additional-order tests have Holm-adjusted $p=0.0050$.
We then preserve every role position and permute only within the same role.
These two role-preserving checks use a separate Holm adjustment.
The additional gains remain positive and significant under this role-preserving
comparison, as reported in Table~\ref{tab:ant-order}. The results associate
scenario with both semantic ordering and primitive macro group transitions
beyond composition and role arrangement.

\begin{table}[t]
\centering
\small
\caption{Aggregate additional sequence cohesion. Observed and reference-mean
gains are percentages, and $\Delta G$ is in percentage points.
Unrestricted-order tests use the six-test Holm adjustment.
Role-preserving checks use a separate two-test adjustment.}
\label{tab:ant-order}
\begin{tabular}{@{}llrrrr@{}}
\toprule
Representation & Reference order & Observed & Reference mean & $\Delta G$ & $p_{\mathrm H}$ \\
\midrule
Directed area & Unrestricted & 6.74 & 0.97 & 5.77 & 0.0050 \\
Transition residual & Unrestricted & 5.70 & 0.46 & 5.24 & 0.0050 \\
Directed area & Role-preserving & 6.74 & 3.26 & 3.48 & 0.0020 \\
Transition residual & Role-preserving & 5.70 & 1.42 & 4.28 & 0.0020 \\
\bottomrule
\end{tabular}
\end{table}
\section{Experiment Details}
\subsection{Baseline Descriptions}
\label{app:baseline-descriptions}

We evaluate five encrypted traffic classification baselines and eight website fingerprinting baselines.

\subsubsection{Encrypted Traffic Classification}
\label{app:encrypted-traffic-baselines}
\begin{itemize}
  \setlength{\itemsep}{0pt}
  \setlength{\parskip}{0pt}
  \setlength{\parsep}{0pt}
  \item \textbf{FlowLens}~\citep{barradas2021flowlens}: Computes compact, application-dependent flow markers from packet-size and inter-packet-time distributions for line-rate classification on programmable switches.
  \item \textbf{YaTC}~\citep{yu2023yatc}: Formats raw packet bytes as a hierarchical flow matrix and applies packet-level and flow-level attention with masked-autoencoder pretraining.
  \item \textbf{NetMamba}~\citep{wang2024netmamba}: Applies a Mamba state-space model to capture long-range sequential traffic dependencies in linear time.
  \item \textbf{TFusion}~\citep{tfusion2025}: Fuses per-packet sequential features with per-flow and per-host statistical features before training lightweight downstream classifiers.
  \item \textbf{TrafficFormer}~\citep{trafficformer2025}: Represents traffic as packet or burst tokens and uses a Transformer encoder to learn contextual relationships.
\end{itemize}

\subsubsection{Website Fingerprinting}
\label{app:website-fingerprinting-baselines}
\begin{itemize}
  \setlength{\itemsep}{0pt}
  \setlength{\parskip}{0pt}
  \setlength{\parsep}{0pt}
  \item \textbf{Deep Fingerprinting (DF)}~\citep{sirinam2018deepfingerprinting}: A deep CNN that classifies packet-direction sequences by capturing local and hierarchical patterns.
  \item \textbf{AWF}~\citep{rimmer2018automated}: An early deep-learning baseline extracting discriminative representations directly from packet-level traffic sequences.
  \item \textbf{Tik-Tok}~\citep{rahman2020tiktok}: Leverages inter-packet timing patterns alongside packet ordering and direction to capture temporal characteristics.
  \item \textbf{Var-CNN}~\citep{bhat2019varcnn}: Uses parallel convolutional branches to model complementary views such as direction and timing for improved data efficiency.
  \item \textbf{NetCLR}~\citep{bahramali2023netclr}: Employs contrastive learning and trace augmentation to learn representations that are robust to network variations.
  \item \textbf{TMWF}~\citep{jin2023tmwf}: Treats multi-tab recognition as ordered-set prediction and uses single-tab queries to extract website features from different trace positions.
  \item \textbf{ARES}~\citep{deng2023ares}: Formulates multi-tab fingerprinting as multi-label classification, extracts local patterns from multi-level traffic aggregation features, and models their correlations with improved self-attention.
  \item \textbf{BAPM}~\citep{guan2021bapm}: Uses block-level attention to profile mixed traces for multi-tab website fingerprinting.
\end{itemize}

\subsection{Experiment Configuration}
\label{app:experiment-configuration}

\paragraph{Data splits.}
All benchmarks share fixed episode splits randomly assigned within each task, and behavior primitive segments inherit their source episode's split. The episode test set contains 209 episodes: 189 benign and 20 malicious, with scenario counts of 35, 72, 60, 22, and 20 in the order listed in Section~\ref{sec:task_definition}. The corrected primitive split counts are 35,400 training, 2,529 validation, and 2,101 test segments before oversampling; the reported primitive evaluation covers 2,101 test segments.

Table~\ref{tab:experiment-configuration} summarizes the configurations used for risk, scenario, and primitive identification. Risk and scenario identification use two and five classes, respectively, with predictions evaluated at the episode level. Primitive identification uses 47 macro groups and is evaluated at the behavior primitive segment level.

\begin{center}
\centering
\captionof{table}{Main experiment configurations. R, S, and P denote risk, scenario, and primitive identification. LR and WD denote learning rate and weight decay. DIR denotes packet directions, DT denotes signed relative timestamps, and DT2 denotes separate direction and interarrival-time channels. Lengths refer to model inputs. RF denotes random forest; $n$ and $d$ denote its number of trees and maximum depth.}
\label{tab:experiment-configuration}
\small
\setlength{\tabcolsep}{4pt}
\resizebox{\textwidth}{!}{%
\begin{tabular}{lllrlcrl}
\toprule
Baseline & Task & Input / length & Batch & LR & Optimizer & Epochs & Additional settings \\
\midrule
FlowLens & R, S & 4 flow statistics & -- & -- & RF & -- & $n=100$, unrestricted depth \\
YaTC & R, S & $40\times40$ bytes & 256 & $2.00\times10^{-3}$ & AdamW & 200 & WD=0.05; warmup=20 epochs \\
NetMamba & R & $40\times40$ bytes & 128 & $1.335\times10^{-3}$ & AdamW & 40 & WD=0.05; warmup=20 epochs \\
NetMamba & S & $40\times40$ bytes & 3072 & $1.068\times10^{-2}$ & AdamW & 40 & WD=0.05; warmup=20 epochs \\
TFusion & R & Segment length 200 & 200 & $2.259\times10^{-5}$ & Adam & 15 & WD=$3.293\times10^{-5}$; RF: $n=200$, $d=10$ \\
TFusion & S & Segment length 100 & -- & -- & RF & -- & Frozen encoder; $n=20$, $d=50$ \\
TrafficFormer & R & 256 tokens & 32 & $2.033\times10^{-5}$ & AdamW & 4 & Dropout=0.455; warmup=5.18\% \\
TrafficFormer & S & 320 tokens & 96 & $4.543\times10^{-5}$ & AdamW & 4 & Dropout=0.5; warmup=10\% \\
\midrule
DF & R, S, P & DIR / 5,000 & 128 & $2\times10^{-3}$ & Adamax & 30 & WD=0 \\
AWF & R, S, P & DIR / 3,000 & 256 & $8\times10^{-4}$ & RMSprop & 30 & WD=0 \\
TikTok & R, S, P & DT / 5,000 & 128 & $2\times10^{-3}$ & Adamax & 30 & WD=0 \\
VarCNN & R, S, P & DT2 / 5,000 & 50 & $10^{-3}$ & Adam & 30 & WD=0 \\
NetCLR & R, S, P & DIR / 5,000 & 256 & $3\times10^{-4}$ & Adam & 30 & WD=0 \\
TMWF & R, S, P & DIR / 30,720 & 80 & $5\times10^{-4}$ & Adam & 30 & WD=0 \\
ARES & R & MTAF / $8\times8000$ & 128 & $5\times10^{-4}$ & AdamW & 40 & WD=0.01; 20\,ms bins \\
ARES & S, P & MTAF / $8\times8000$ & 512 & $2\times10^{-3}$ & AdamW & 30 & WD=0.01; 20\,ms bins \\
BAPM & R, S, P & DIR / 8,500 & 128 & $5\times10^{-4}$ & Adam & 30 & WD=0 \\
\bottomrule
\end{tabular}%
}

\vspace{3pt}
\begin{minipage}{0.98\textwidth}
\footnotesize
For YaTC and NetMamba, LR is the base absolute learning rate before warmup and decay. TFusion's risk batch size and epoch count apply to encoder training. TrafficFormer uses AdamW for risk identification; its four epochs indicate the training budget. Dashes denote inapplicable settings. Numerical hyperparameters are rounded where necessary.
\end{minipage}
\end{center}

The eight WF baselines use packet sequences, truncated or zero-padded to the lengths in Table~\ref{tab:experiment-configuration}. ARES converts the first 10,000 packets into eight-channel temporal features using 20\,ms bins. TMWF uses a model input length of 30,720, although its risk inputs retain only the first 10,000 packets before padding. We train the WF models with cross-entropy loss and fixed learning rates. All checkpoints are selected by validation accuracy. For primitive identification, training classes containing fewer than 200 segments are oversampled to 200 by random sampling with replacement within each class; validation and test sets retain their original distributions.

FlowLens operates on packet count, total bytes, median interarrival time, and flow duration. YaTC and NetMamba are fine-tuned from pretrained checkpoints, with weight decay 0.05, layer-wise learning-rate decay 0.75, and label smoothing 0.1. TFusion's risk encoder is trained contrastively on benign flows, followed by a random forest with balanced class weights and a minimum leaf size of five. For scenario identification, TFusion freezes an encoder previously trained on benign ANT flows and trains only the five-class random forest. TrafficFormer uses task-specific sequence lengths, learning rates, and dropout rates, as listed in the table.

For episode-level evaluation of encrypted traffic classification baselines, we select the aggregation strategy with the highest validation-set accuracy among \texttt{mean\_soft}, \texttt{majority\_vote}, \texttt{confidence\_weighted\_mean}, and \texttt{max\_confidence}. For scenario identification, flow-level class probabilities are averaged within each episode before taking the argmax. Risk identification uses the same aggregation for FlowLens and YaTC, and majority voting for NetMamba, TFusion, and TrafficFormer.

\end{document}